\documentclass[10pt,twocolumn,letterpaper]{article}

\usepackage[pagenumbers]{config/cvpr}
\usepackage{xspace}
\usepackage{graphicx}
\usepackage[dvipsnames]{xcolor}
\usepackage{comment}
\usepackage{amsmath,amssymb}
\usepackage{color}
\usepackage{microtype}
\usepackage[accsupp]{axessibility}
\usepackage{caption}
\usepackage{subcaption}
\usepackage{booktabs}
\usepackage{soul}
\usepackage[pagebackref,breaklinks,colorlinks]{hyperref}

\newcommand{\colorRef}[1]{\textcolor{red}{#1}}
\usepackage[capitalize]{cleveref}
\crefname{figure}{\colorRef{Fig.}}{\colorRef{Figs.}}
\Crefname{figure}{\colorRef{Figure}}{\colorRef{Figures}}
\crefname{section}{\colorRef{Sec.}}{\colorRef{Secs.}}
\Crefname{section}{\colorRef{Section}}{\colorRef{Sections}}
\Crefname{table}{\colorRef{Table}}{\colorRef{Tables}}
\crefname{table}{\colorRef{Tab.}}{\colorRef{Tabs.}}

\newcommand{\ourTitle}{Detecting Human-Object Contact in Images}

\usepackage{acronym}
\acrodef{hoi}[HOI]{Human-Object Interaction}
\acrodef{hot}[HOT]{Human-Object conTact}
\acrodef{amt}[AMT]{Amazon Mechanical Turk}
\acrodef{hps}[HPS]{Human Pose and Shape}
\acrodef{iou}[IoU]{Intersection-over-Union}

\newcommand{\colorPOP}{black}

\newcommand{\supmat}{\textcolor{magenta}{Appx.}\xspace}

\newcommand{\modelname}{\textcolor{\colorPOP}{\texorpdfstring{\acs{hot}}{}}\xspace}
\newcommand{\modelnameLong}{\textcolor{\colorPOP}{``\aclu{hot}''}\xspace}

\newcommand{\setAnnot}{\textcolor{\colorPOP}{``\hot-Annotated''}\xspace}
\newcommand{\setGener}{\textcolor{\colorPOP}{``\hot-Generated''}\xspace}

\newcommand{\setTotalIMG}{\textcolor{\colorPOP}{$35,750$}\xspace}

\newcommand{\setTotalAREAS}{\textcolor{\colorPOP}{$163,534$}\xspace}

\newcommand{\prox}{\mbox{\textcolor{\colorPOP}{PROX}}\xspace}
\newcommand{\vcoco}{\mbox{\textcolor{\colorPOP}{V-COCO}}\xspace}
\newcommand{\hake}{\mbox{\textcolor{\colorPOP}{HAKE}}\xspace}
\newcommand{\wnp}{\mbox{\textcolor{\colorPOP}{Watch-n-Patch}}\xspace}
\newcommand{\hot}{\mbox{\modelname}\xspace}

\usepackage{lipsum}

\newcommand{\TODO}[1]{\xspace{\textcolor{red}{#1}}\xspace}

\renewcommand{\etc}{etc\xspace}
\renewcommand{\etal}{et al.\xspace}
\renewcommand{\ie}{i.e.\xspace}
\renewcommand{\eg}{e.g.\xspace}

\newcommand{\dummytext}[1]{
\newcount\zz
\loop
\TODO{Dummy one-line filler text.}
\TODO{Dummy one-line filler text.}
\advance\zz1
\ifnum\zz<#1
\repeat
}

\newcommand{\zheading}[1]{\textbf{#1:}}
\newcommand{\qheading}[1]{\noindent\textbf{#1:}}

\newcommand{\twoD}{2D\xspace}
\newcommand{\threeD}{3D\xspace}

\newcommand{\inthewild}{{in-the-wild}\xspace}

\usepackage{cuted}
\usepackage{bbm}
\usepackage{enumitem}
\usepackage{tabularx,multirow,array,makecell}
\usepackage{wrapfig}

\usepackage[normalem]{ulem} %

\newcommand{\mocap}{\mbox{MoCap}\xspace}

\newcommand{\smpl}{\mbox{SMPL}\xspace}
\newcommand{\smplx}{\mbox{SMPL-X}\xspace}
\newcommand{\smplX}{\smplx}

\newcommand{\groundtruth}{{ground-truth}\xspace}

\usepackage{balance}

\newcommand{\websiteURL}{\mbox{\url{https://hot.is.tue.mpg.de}}}

\begin{document}
\title{\ourTitle}

\author{
Yixin Chen$^{1 \dagger}$   \quad
Sai Kumar Dwivedi$^{2}$    \quad
Michael J. Black$^{2}$     \quad
Dimitrios Tzionas$^{3}$\\
\small
\hspace{-2.2 em}
$^{1}$Beijing Institute of General Artificial Intelligence, China~ 
\small
\hspace{+4.6 em}
$^{\dagger}$Work done while interning at MPI-IS$^{2}$ 
\\
\small
$^{2}$Max Planck Institute for Intelligent Systems, T{\"u}bingen, Germany~ 
\hspace{+1.0 em}
$^{3}$University of Amsterdam, the Netherlands
}

\maketitle

\begin{abstract}
Humans constantly contact objects to move and perform tasks. 
Thus, detecting human-object contact is important for building human-centered artificial intelligence.
However, there exists no robust method to detect contact between the body and the scene from an image, and there exists no dataset to learn such a detector.
We fill this gap with \modelname (\modelnameLong), a new dataset of human-object contacts in images. 
To build \hot, we use two data sources:
(1)     We use the \prox dataset of \threeD human meshes moving in \threeD scenes, and automatically annotate \twoD image areas for contact via \threeD mesh proximity and projection.
(2)      We use the \vcoco, \hake and Watch-n-Patch datasets, and ask trained annotators to draw polygons around the \twoD image areas where contact takes place. 
We also annotate the involved body part of the human body.
We use our \modelname dataset to train a new contact detector, which takes a single color image as input, and outputs \twoD contact heatmaps as well as the body-part labels that are in contact.
This is a new and challenging task, that extends current foot-ground or hand-object contact detectors to the full generality of the whole body.
The detector uses a part-attention branch to guide contact estimation through the context of the surrounding body parts and scene. 
We evaluate our detector extensively, and quantitative results show that our model outperforms baselines, and that all components contribute to better performance. 
Results on images from an online repository show reasonable detections and generalizability.
Our \modelname data and model are available for research at \websiteURL.
\end{abstract}

\section{Introduction}
\label{sec:intro}

Contact is an important part of people's everyday lives. 
We constantly contact objects to move and perform tasks. 
We walk by contacting the ground with our feet, 
we sit by contacting chairs with our buttocks, hips and back, 
we grasp and manipulate tools by contacting them with our hands. 
Therefore, estimating contact between humans and objects is useful for human-centered AI, especially for applications such as AR/VR~\cite{argall2009survey,grover2021pipeline,kumar2015mujoco,lipton2017baxter}, 
activity recognition~\cite{jia2020lemma,rai2021home,savva2016pigraphs}, 
affordance detection~\cite{fouhey2012people,koppula2014physically,nagarajan2020learning,zhu2016inferring}, 
fine-grained human-object interaction detection~\cite{li2020pastanet,qi2018learning,wang2019deep,xu2019learning}, 
imitation learning~\cite{radosavovic2020state,thobbi2010imitation,zhang2018deep}, 
populating scenes with avatars~\cite{hassan2021populating,zhang2020place,zhang2020generating}, 
and sanitization of spaces and objects.

In contrast to off-the-shelf detectors for segmenting humans in images, or estimating their \twoD joints or \threeD shape and pose,  there exists no general detector of contact.
Some work exists for detecting part-specific contact, \eg, hand-object ~\cite{narasimhaswamy2020detecting,shan2020understanding} or foot-ground~\cite{rempe2020contact,tripathi2023ipman} contact, while other work estimates contact only in constrained environments~\cite{shimada2022hulc,huang2022capturing} with limited generalization.
What we need, instead, is a contact detector for the \emph{entire body} that estimates detailed, body-part-related, contact \emph{maps} in arbitrary images. 
To train this, we need data, but no suitable dataset exists at the moment. 
We address these limitations  with a novel dataset and model for detecting contact between whole-body humans and objects in color images taken in the wild.

\newcommand{\teaserCaption}{
        Our contact detector, trained on \modelname (\modelnameLong) dataset, estimates contact between humans and scenes from an image taken in the wild.
        Contact is important for interacting humans, yet, standard \inthewild datasets unfortunately lack such information. 
        Our contact dataset and detector are a step towards providing this in the wild. 
        Images are from \href{https://www.pexels.com}{pexels.com}. 
}

\begin{figure}[t!]
	\centering
	\includegraphics[width=\linewidth]{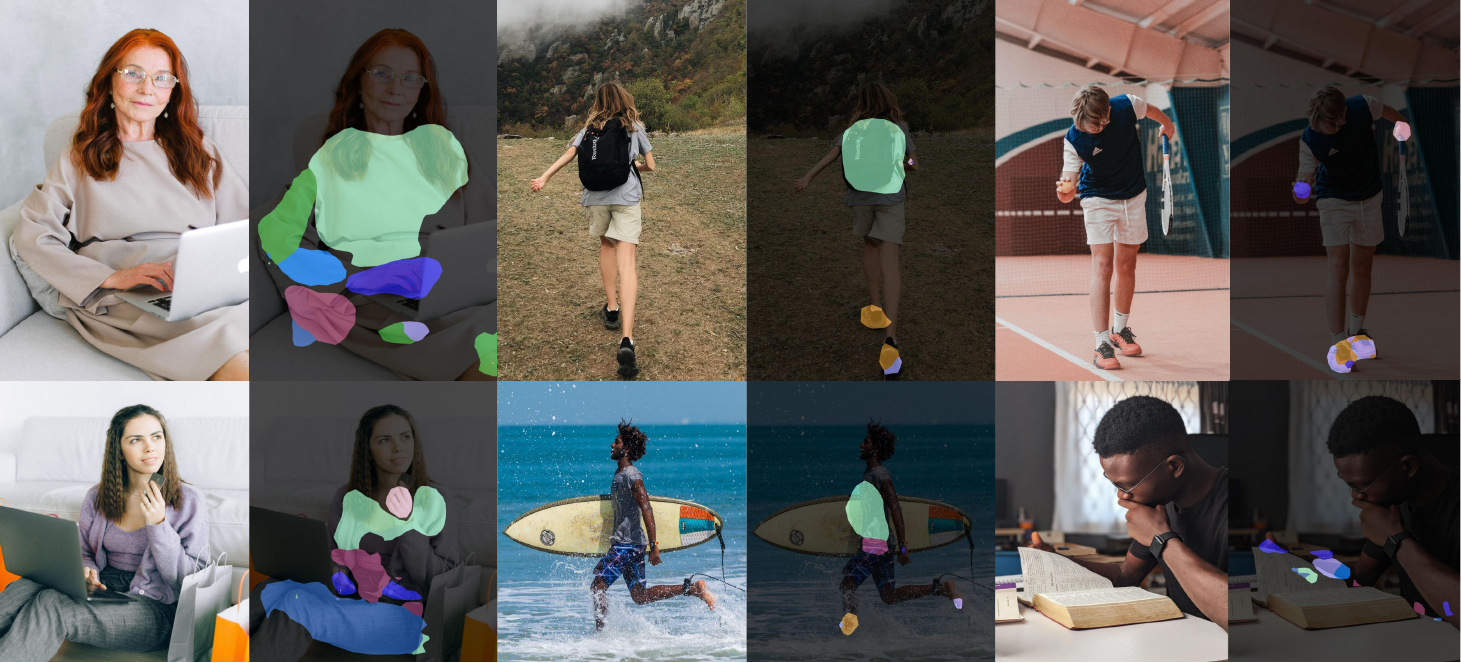}
	\caption{\teaserCaption}
	\label{fig:teaser}
\end{figure}

Annotating contact is challenging, as contact areas are ipso facto occluded. 
Think of a person standing on the floor; the sole of the shoe, and the floor area it contacts, can not be observed. 
A naive approach is to instrument a human with contact sensors, however, this is intrusive, cumbersome to set up and does not scale. 
Instead, we use two alternative data sources, with different but complementary properties: 
\mbox{(1) 
We use the} \prox~\cite{hassan2019resolving} dataset, which has pseudo \groundtruth \threeD human meshes for real humans moving in \threeD scanned scenes. 
We \emph{automatically} annotate contact areas, by computing  the proximity between the \threeD meshes.
\mbox{(2) 
We use the} \vcoco~\cite{gupta2015visual}, \hake~\cite{li2020pastanet}, and \wnp~\cite{wu2015watch} datasets, which contain images taken in the wild. 
We then hire professional annotators, and train them to annotate contact areas as \twoD polygons in images. 
Although \emph{manual} annotation is only approximate, \twoD annotations are important because they allow scaling to large, varied, and natural datasets.  This improves generalization.
Note that in both cases we also annotate the body part that is involved in contact, corresponding to the body parts of the \mbox{SMPL(-X)} \cite{SMPL:2015,smplifyPP} human model. 

\begin{figure}
\begin{subfigure}[t]{\linewidth}
		\includegraphics[width=0.497\linewidth,trim={0cm 0cm 0cm 0cm},clip]{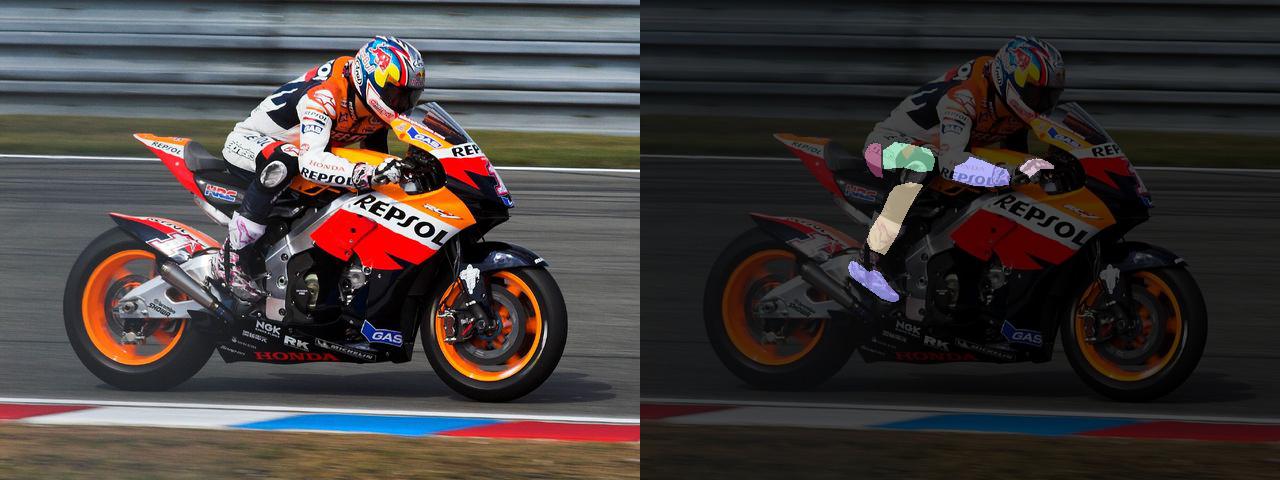}\hfill
		\includegraphics[width=0.497\linewidth,trim={0cm 0cm 0cm 0cm},clip]{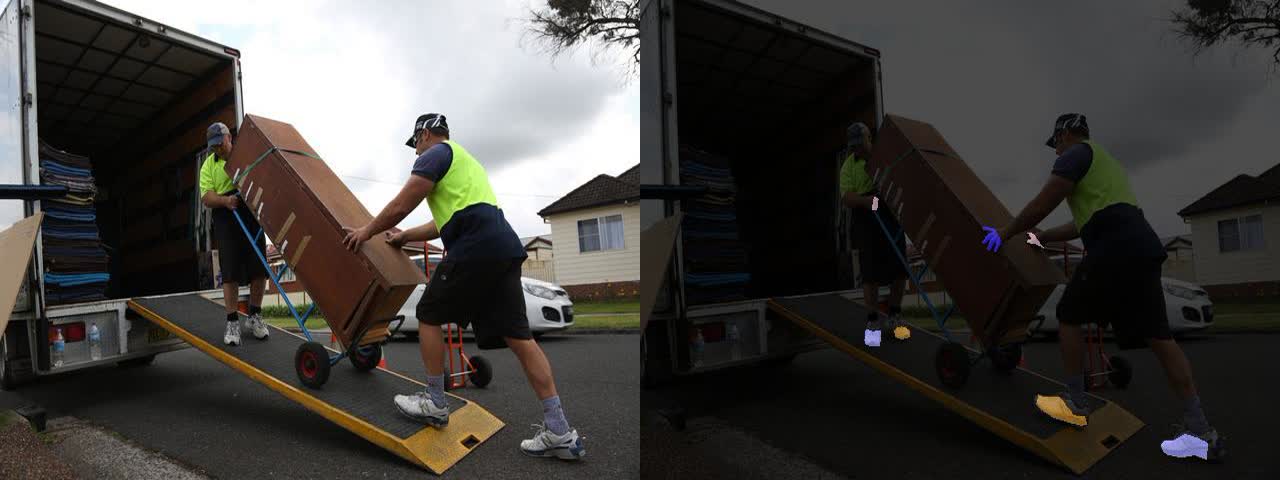}
\\
		\includegraphics[width=0.497\linewidth,trim={0cm 0cm 0cm 0cm},clip]{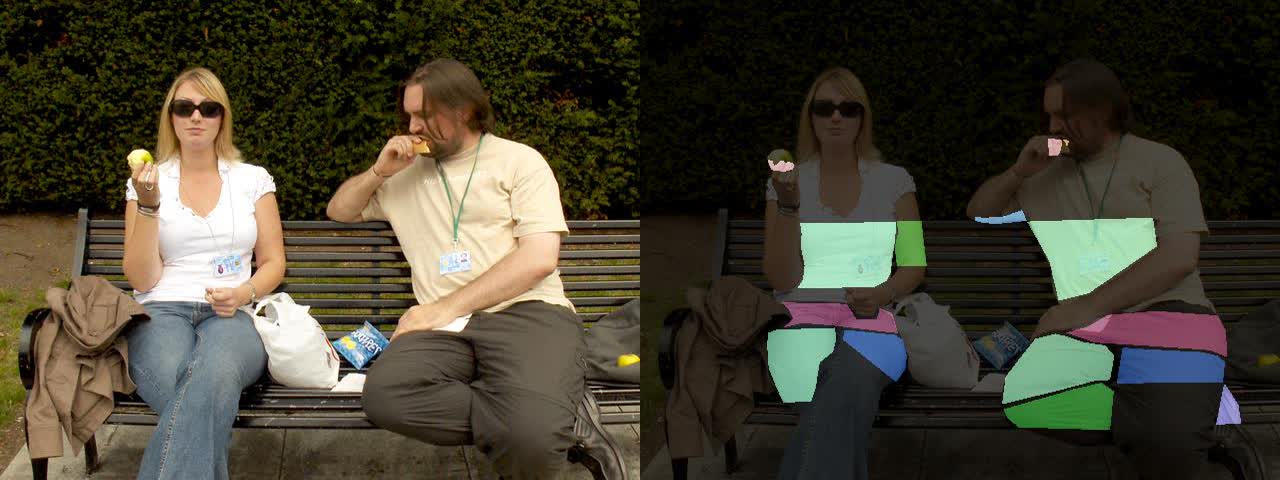}\hfill
		\includegraphics[width=0.497\linewidth,trim={0cm 0cm 0cm 0cm},clip]{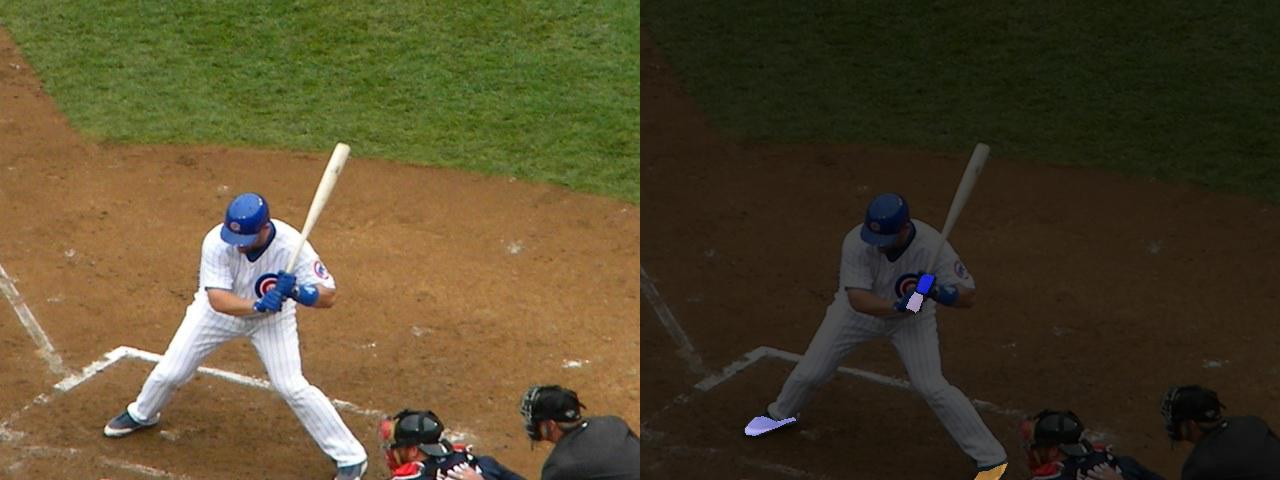}
\caption{\setAnnot examples.}
\label{fig:anno_ex}
\end{subfigure}%
\\
\begin{subfigure}[t]{0.494\linewidth}
\vspace{-2.85cm}
		\includegraphics[width=\linewidth,trim={0cm 0.1cm 0cm 0.1cm},clip]{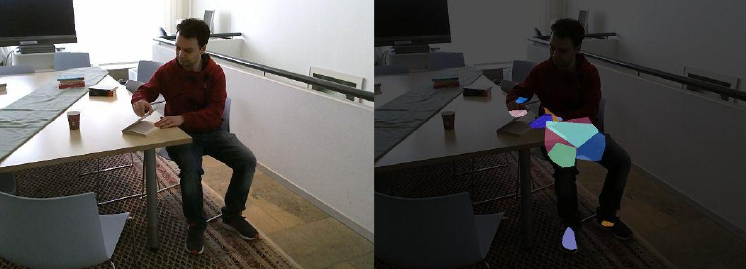}\\
		\includegraphics[width=\linewidth,trim={0cm 0.1cm 0cm 0.1cm},clip]{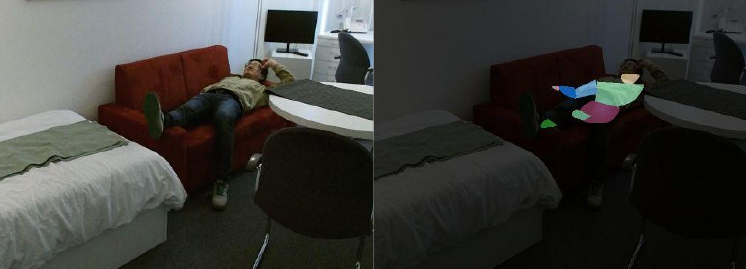}
		\caption{\setGener examples.}
		\label{fig:gen_ex}
\end{subfigure}%
\hfill
\begin{subfigure}[t]{0.504\linewidth}
		\includegraphics[width=\linewidth,trim={0cm 0cm 0cm 0cm},clip]{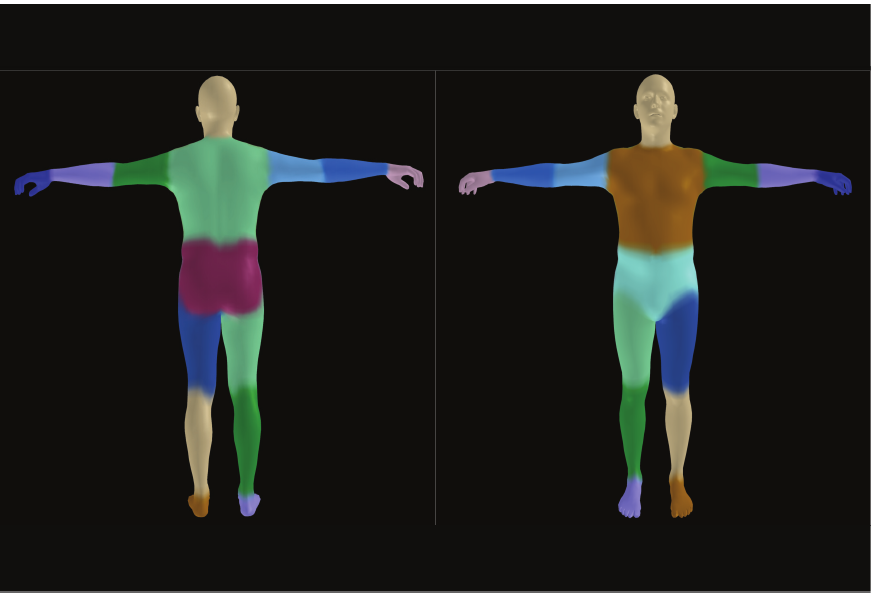}
		\caption{Human body parts of \smplX.}
		\label{fig:human_part}
\end{subfigure}%
\caption{
            Images and contact annotations for our \hot dataset. 
            We show examples for both its parts, \ie, 
            \setGener (\cref{sec:SetGener})
            and 
            \setAnnot (\cref{sec:SetAnnot}).
            Contact annotations include the involved body part (c), shown color coded on a \smplX mesh. 
}
\label{fig:dataset}
\end{figure}

We thus present \hot (\modelnameLong), a new dataset of images with human-object contact; see examples in \cref{fig:dataset}. 
The first  part of \hot, called \setGener (\cref{fig:dataset}\textcolor{red}{b}), has automatic annotations, but lacks variety for human subjects and scenes. 
The second part, called \setAnnot (\cref{fig:dataset}\textcolor{red}{a}), has manual annotations, but has a huge variety of people, scenes and interactions. 
\hot has \setTotalIMG images with \setTotalAREAS contact annotations. 

We then train a new contact detector on our \hot dataset. 
Given a single color image as input, we want to know, if contact takes place in the image, the area in which it occurs, as well as the body part that is involved. 
Specifically, we detect \twoD heatmaps in an image, encoding the contact location and likelihood, and classify each pixel in contact to one of \mbox{SMPL(-X)}'s body parts.  
However, training directly with \hot annotations leads to ``bleeding'' heatmaps and false detections. 
We observe that humans reason about contact by looking at body parts and their proximity to objects in their local vicinity. 
Therefore, we use a body-part-driven attention module that significantly boosts performance.

We evaluate our detector on withheld parts of the \hot dataset. 
Quantitative evaluation and ablation studies show that our model outperforms the baselines, and that all components contribute to detection performance. 
Our body-part attention module is the key component; a visual analysis shows that it attends to meaningful image locations, \ie, on body parts and their vicinity. 
Qualitative results show reasonable detections on \inthewild images. 
By applying our detector on datasets unseen during training, we show that the model generalizes reasonably well; see \cref{fig:teaser}. 
Then, we show that our general-purpose full-body contact detector performs on par with existing part-specific contact detectors for the foot \cite{rempe2020contact} or hand \cite{narasimhaswamy2020detecting}, 
meaning it could serve as  a drop-in replacement for these.
Moreover, we show that our contact detector helps contact-driven \threeD human pose estimation on \prox data \cite{hassan2019resolving}.
Finally, we show that our \hot dataset helps a state-of-the-art (SOTA) 3D body-scene contact estimator \cite{huang2022capturing} generalize to \inthewild images. 

In summary, \hot takes a step towards automatic contact detection between humans and objects in color images and our contributions are three-fold: 
\mbox{(1)     We introduce} the  task of full-body human-object contact detection in images. 
\mbox{(2)     To facilitate} machine learning for this, we introduce the \hot dataset with \twoD contact area heatmaps and the associated human part labels as annotations, using both auto-generated and manual annotations. 
\mbox{(3)     We develop} a new contact detector that incorporates a body-part attention module. 
Experiments and ablations show the benefits of the proposed model and its components.
Our data and code are available at \websiteURL.

\section{Related Work}
\label{sec:related}

\paragraph{Contact Modeling.}
Prior work on modeling the contact between humans and the world can be classified as either ``body-object" or ``hand-object."

\zheading{Body-object contact}
Several works model the contact between the human body and object~\cite{clever2020bodies,xie2022chore,yi2022mime,dai2023sloper4d,huang2023scenediff,jiang2022chairs}.
Li \etal~\cite{li2019estimating} reconstruct the \threeD motion of a person interacting with an object by estimating the \threeD pose of the person and object, the joint-level contact, and forces and torques actuated by the human limbs. 
Rempe \etal~\cite{rempe2021humor,rempe2020contact} estimate joint-level foot-ground contact from a video, and use it to constrain the human pose with trajectory optimization. 
More recently, \cite{huang2022capturing,shimada2022hulc} propose to directly estimate 3D body-scene contact, but the lack of data variety limits the model's generalization to \inthewild scenarios, even when a 3D scene is used as additional input \cite{shimada2022hulc}.
Others use \ac{hoi} relationships to reconstruct~\cite{chen2019holistic++,weng2021holistic,li2022mocapdeform,bhatnagar2022behave} or generate~\cite{zhang2020place,zhang2020generating,hassan2021populating} \threeD human and object pose by encouraging contact and penalizing inter-penetrations. 
Prior work uses contact information as prior knowledge, but it is often oversimplified as (1) body-ground contact at the skeleton-joint level, (2) hand-object contact at a rough bounding-box level, or (3) manually-annotated contacts of other human parts in constrained environments. 
In this paper, we seek to automatically estimate contact heatmaps for the whole body in a bottom-up manner directly from a \twoD image. 
We also predict the associated human body part label, which provides a more systematic understanding of human-object contact. 

\zheading{Hand-object contact}
People interact with objects using their hands, so contact plays an important role in hand-object interaction and grasping. 
Contact information is often captured as a byproduct in grasp datasets~\cite{brahmbhatt2020contactpose,liu2021semi,taheri2020grab,fan2023arctic} through hand-object proximity or thermal information. 
Hand-object grasp reconstruction also employs contact to refine the hand and object pose estimation~\cite{cao2021reconstructing,grady2021contactopt,hasson2019learning,wang2022visual,tse2022s}. 
In addition, some works \cite{narasimhaswamy2020detecting,shan2020understanding,zhang2022fine} detect hands and classify their physical contact state into self-contact, person-person contact, and person-object contact. 
Although they consider the relationship between hands and other objects in the scene, they detect only a rough bounding box or boundary for the hand, instead of a finer-grained contact area. 
In this paper, we take a step further to estimate general-purpose full-body contact from \twoD images at a finer scale. 

\paragraph{\texorpdfstring{\acf{hoi}}{}.}
The goal of \ac{hoi} understanding~\cite{qi2018learning,wang2019deep,xu2019learning} is to infer the interaction relationships between humans and objects. 
While both humans and objects are located in the image, often in the form of \twoD bounding boxes, the literature rarely focuses on how the interaction takes place, whether the interaction requires contact, and which human part is involved in the contact.
This limits the applicability of current \ac{hoi} detections for downstream scene understanding tasks.
Recently, Li \etal~\cite{li2020pastanet} provide more detailed body-part state annotations in the context of \ac{hoi}, and offer action labels (\eg, hold, paddle) and the involved human body part (\eg, hand, foot). 
However, they do not annotate \twoD contact areas in images, and their predefined human parts are not fine-grained enough to capture everyday \ac{hoi} scenarios.
In contrast, our new dataset contains \twoD contact areas that are also associated with the involved human body parts following the part segmentation of the popular \mbox{SMPL(-X)}~\cite{SMPL:2015,smplifyPP} statistical \threeD body model. 

\paragraph{Affordance Learning.}
Contact and \ac{hoi} are closely related to object affordances, which reflect the functional aspects of an object. 
Recent work explores object affordance learning from human actions and object manipulation~\cite{fouhey2012people,koppula2014physically,nagarajan2020learning,zhu2016inferring}. 
More specifically, Fang \etal~\cite{fang2018demo2vec} and Nagarajan \etal~\cite{nagarajan2019grounded} learn to predict the interaction region with the corresponding action label on a target object from human demonstration videos. 
Savva \etal~\cite{savva2014scenegrok,savva2016pigraphs} capture physical contact and visual attention links between \threeD geometry and human body parts from RGB-D videos. 
Deng \etal~\cite{deng20213d} collect a \threeD visual affordance dataset with potential interaction areas on \threeD objects for various actions. 
Affordance learning is object-centric; it 
does not capture much about the human actor. 
In contrast, detecting interaction areas (\eg contact heatmaps) reflects how people interact with objects, and considers both the human and the object. 

\section{\texorpdfstring{\acf{hot}}{} Dataset}
\label{sec:dataset}

To facilitate research in contact estimation, we introduce \hot, a new dataset with \twoD contact areas and the associated human part labels as annotations.
Annotating and detecting contact in images is challenging, as contact depends on the scene and its objects, the humans, the camera view and the occlusions arising from all these factors. 
To create a well-varied dataset, we collect images from two different sources and gather contact annotations. 
Below we discuss the creation of \hot and provide a comprehensive analysis. 

\begin{figure*}[t!]
\centering
\vspace{-1.0 em}
\includegraphics[trim=000mm 000mm 000mm 000mm, clip=false, width=1.00 \linewidth]{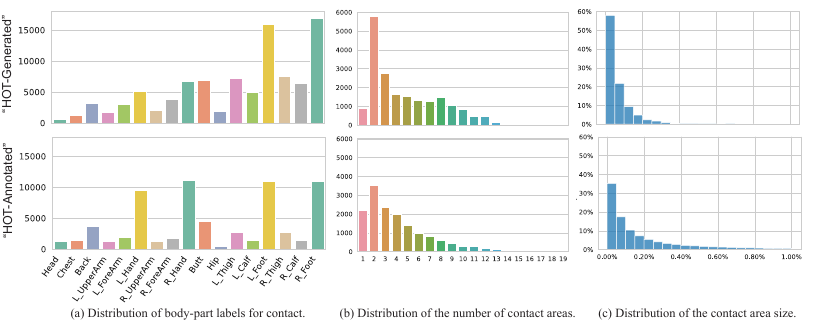}
    \vspace{-2.0 em}
    \caption{
                Data statistics. 
                \textbf{(a)}     
                Number of contact areas (Y-axis) for each body part (X-axis). 
                \textbf{(b)}     
                Number of images (Y-axis) with a certain number of contact areas (X-axis). 
                \textbf{(c)}     
                Percentage of contact areas (Y-axis) that occupy a certain percentage of image pixels (X-axis). 
    }
    \label{fig:statistics}
\end{figure*}

\subsection{Data Sources}

\zheading{\setGener}
First we collect data from the \prox dataset \cite{hassan2019resolving}, which contains people reconstructed as \threeD~\smplX~\cite{smplifyPP} meshes interacting with static \threeD scenes; this involves actions like sitting, walking, lying down, \etc. 
Recent work \cite{rempe2021humor,zhang2021lemo} improves on the quality of reconstructed meshes in \prox, facilitating the automatic generation of contact heatmaps by simply using \threeD proximity metrics between the \threeD human mesh and the static \threeD scene mesh. 
We sub-sample frames from the ``qualitative set'' of \prox, and form the \setGener part of \hot.

\zheading{\setAnnot}
Another source for images with human-object contact is \ac{hoi} datasets like \vcoco~\cite{gupta2015visual} and \hake~\cite{li2020pastanet}. 
As they are collected from Flickr, these datasets contain very diverse \ac{hoi} interactions in complex and cluttered scenes. 
Existing \ac{hoi} datasets contain activity labels and bounding boxes for humans and objects, but boxes are too coarse for understanding contact. 
Thus, we select a subset from the \vcoco~\cite{gupta2015visual} and \hake~\cite{li2020pastanet} datasets and use these to gather new contact annotations. 
To keep the task tractable, we first remove images with indirect human-object interaction, heavily cropped humans, motion blur, distortion or extreme lighting conditions. 
Other interesting datasets are indoor action recognition datasets like \wnp~\cite{wu2015watch} that contain several daily activities like ``fetch-from-fridge'', ``put book back'', \etc. 
We sample image frames from video clips where human subjects and objects are clearly visible. 
We then combine images selected from \vcoco~\cite{gupta2015visual}, \hake~\cite{li2020pastanet} and \wnp~\cite{wu2015watch}, and form the \setAnnot part of \hot. 

\subsection{Contact Generation for \setGener}
\label{sec:SetGener}

The \prox dataset \cite{hassan2019resolving} captures people interacting with static scenes. 
\prox represents the human pose and shape with the \smplX~\cite{smplifyPP} body model, which captures the body surface  including the hands and face. 
The \smplX model represents the human body with pose parameters $\theta$, and shape parameters $\beta$, and outputs a posed \threeD mesh, $\mathcal{M}_b \in \mathbb{R}^{10475\times3}$. 
Each vertex, $v \in \mathbb{R}^{3}$, has a surface normal, $n^{v}$, and a human part label, $c$, associated with it. 
We divide the \smplX model into $17$ parts $c_i$, with $i \in \{1, 2, ..., 17\}$. 
The body parts are based on the original part segmentation of \smplX and, for simplicity, we merge  parts that  human annotators cannot easily differentiate; \eg, parts of the back across the spine.
See \cref{fig:human_part} for the color-coded segmentation of \smplX into corresponding body-part labels.

For each frame, given the reconstructed \smplX mesh, $\mathcal{M}_b$, and the scene mesh, $\mathcal{M}_s$, we  calculate human-to-scene mesh distances. 
Then, all human vertices with a distance to the scene below a threshold, and with compatible normals to the scene ones, are annotated as contact vertices. 
Finally, for the contact vertices we find the respective triangles on the \threeD body mesh, and render these separately per body part to get dense \twoD contact areas in the image space. 
In this way, we automatically create pseudo ground truth for contact. 
Examples are shown in \cref{fig:gen_ex}.
For more details on the above steps, see \supmat

\subsection{Contact Annotation for \setAnnot}
\label{sec:SetAnnot}

We gather contact annotations for in-the-wild images without paired 3D human and scene meshes. 
Determining contact areas between a human and an object in an image is non-trivial, as contact areas are always occluded. 
While Amazon Mechanical Turk is popular for annotation collection, the diverse background of its annotators complicates the training, annotation, and quality control for our novel task.
Thus, we hired a professional company with annotators that are already trained for similar tasks.
The annotation has two steps: 
(1)     drawing a polygon around the image areas containing human-object contacts, and 
(2)     assigning a human body-part label to it. 
See \cref{fig:anno_ex} for annotation examples.

We take a number of steps to ensure good quality and consistency for the annotations. 
In particular, we first have a \emph{trial annotation} with 3 rounds for 400 images; we provide task instructions, collect annotations, provide feedback to annotators, and iterate. 
Then, during the \emph{real annotation}, 12 people perform the initial annotation, and we have two rounds of quality checks (QC); a different cohort of 4 people perform the first  QC round, and another 2 people perform the second QC round. 
We use semantic-segmentation annotation tools adapted for our task. 
For more details on the annotation protocol, repeatability and quality check, see \supmat

Compared to the automatic annotations of \setGener, the manual ones of \setAnnot are only approximate. 
However, capturing \threeD people in scenes and accurately reconstructing them in \threeD is hard and does not scale. 
Thus, manual \twoD annotations are important because they allow scaling to large, varied, and natural datasets with images taken in the wild. For data-driven models, this helps  improve their generalization and make them robust.

\begin{figure*}[ht!]
	\centering
	\vspace{-1.5 em}
	\includegraphics[width=1.0\linewidth,trim={0cm 0cm 0cm 0cm},clip]{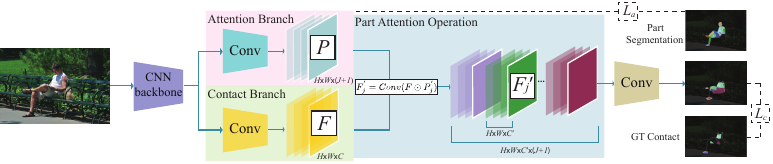}
	\vspace{-2.0 em}
	\caption{
		Architecture for our HOT contact detector.
		Our model takes as input a single color image, and as output it gives \twoD contact heatmaps and 
		a pixel-level classification label for the body part associated with contact. 
    	For a detailed explanation of the model, see \cref{sec:method}. 
	}
	\vspace{-0.2 em}
	\label{fig:architecture}
\end{figure*}

\subsection{\hot Dataset Analysis}

The \hot dataset has a total of $35,287$ images and $162,267$ contact area annotations, along with a body-part label for each area.
Specifically, for \setAnnot we collect 
$5,235$ images and $20,273$ contact areas for \vcoco~\cite{gupta2015visual}, 
$9,522$ images and $45,645$ contact areas for \hake~\cite{li2020pastanet}, and 
$325$  images and   $1,170$ contact areas for the \wnp~\cite{wu2015watch} dataset. 
For \setGener, we auto-generate $95,179$ contact areas in $20,205$ images using the \prox dataset. 
More statistics of \setAnnot and \setGener are shown in \cref{fig:statistics}. 

\noindent
\mbox{\Cref{fig:statistics}{\color{red}a} shows the} distribution of body-part labels for contact. 
We see that \setAnnot has noticeably more contacts than \setGener for both hands. 
The reason is that \prox captures humans interacting with static scenes, \ie, without grasping and moving objects with their hands, while \setAnnot contains a lot of images with interactions between hands and objects. 

\noindent
\mbox{\Cref{fig:statistics}{\color{red}b} shows the} number of contact areas per image. 
We see that \setAnnot has generally more contacts per image than \setGener. 
This is partially because \ac{hoi} datasets contain images of multiple interacting persons, while \prox only has a single person in every image.

\noindent
\mbox{\Cref{fig:statistics}{\color{red}c} shows the} distribution of contact area size. 
We observe that the areas are generally smaller for \setGener than for \setAnnot. 
This is potentially because images in \prox are captured with the camera away from the body to include more scene context, whereas images in \setAnnot are taken in the wild, including close-ups, as well as more object grasps. 

To further analyze the disparities between \setAnnot and \setGener, we had two trained annotators 
annotate 200 random images from \setGener, and we treat these manual annotations as ground truth for evaluating the automatically generated ones. 
The agreement for body-part contact labels is $83.7\%$, while for pixel contact labels, it is $52.4\%$. 
This can be attributed to \prox's noisy \smplX reconstructions due to ambiguities arising from the monocular cameras and strong occlusions. Another contributing factor is the approximate nature of the manual annotations. We provide more discussions in the transfer experiments in \cref{sec:experiments_DataSplits}.

\section{\hot Contact Detector}
\label{sec:method}

To estimate contact areas in images, humans use the global context of the image, but also focus on regions around body parts to examine if there is contact.
Based on these insights, we design our contact detector to extract global features with special attention to human body parts. 

\begin{figure*}[ht]
\includegraphics[width=\linewidth,trim={0cm 0cm 0cm 0cm},clip]{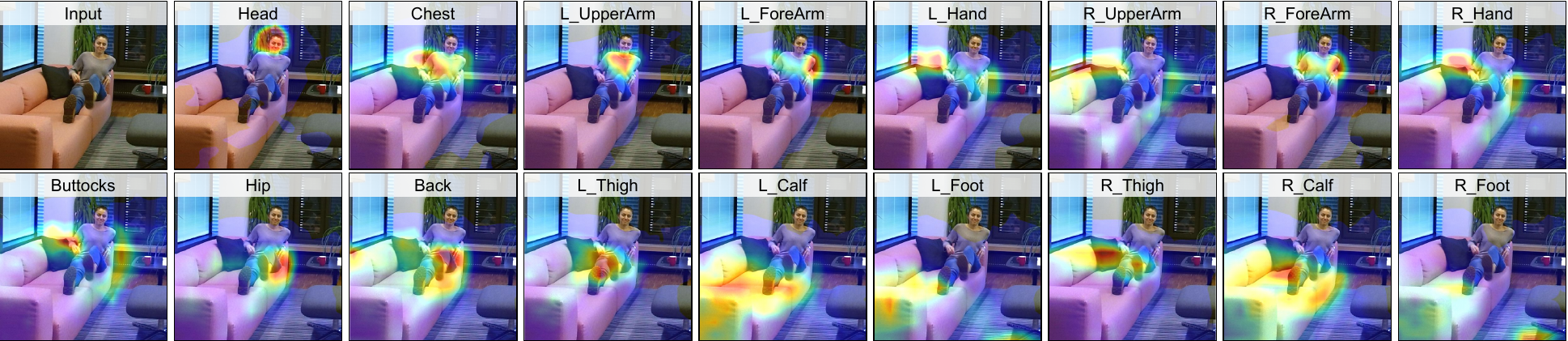}
\vspace{-1.8 em}
\caption{
		Attention visualization. 
		The attention maps from the attention branch, supervised by human part segmentation, 
		guide features to attend to each body-part area, but also to 
		surrounding areas, for reasoning about the nearby scene context and the body-scene interaction. 
}
	\vspace{-1.0 em}
\label{fig:part_attention}
\end{figure*}

\subsection{Architecture}
\Cref{fig:architecture} shows the architecture of our model.
Given an image, we use a CNN backbone to extract image features. 
Then, we use a decoder with two branches: 
an ``\textit{attention branch}'' for inferring an attention mask for each body part and 
a  ``\textit{contact   branch}'' for extracting contact features. 

\qheading{Attention branch}
We denote this  as $P \in \mathbb{R}^{H\times W \times (J+1)}$;
$J$ is the number of body parts, with an extra channel for the background, and 
$H$/$W$ are the feature map's height/width. 
The $j_{th}$ channel $P_j \in \mathbb{R}^{H\times W}$ represents the likelihood that each pixel is associated with contact of the $j_{th}$ body part. 
This guides the model to focus around different human parts in the feature space $F$ of the contact branch. 
By applying a channel-wise softmax normalization $\sigma(.)$ on $P$, we get the attention mask
$P^{'} =  \sigma (P)$, with 
$P^{'} \in \mathbb{R}^{H\times W \times (J+1)} $. 

\qheading{Contact branch}
We denote this as  $F \in \mathbb{R}^{H\times W \times C}$, with the same spatial dimensions $H \times W$ as the attention branch $P$, but with a different number of channels $C=512$. 

\qheading{Part attention operation}
We use the attention mask $P^{'}_{j}$ to extract part-related features from the contact branch $F$: 
\vspace{-0.25 em}
\begin{equation}
    F^{'}_{j} = Conv(F \odot P^{'}_{j})             \text{,}    \quad   \text{with~}
    F^{'}_{j} \in \mathbb{R}^{H\times W \times C'}  \text{,}
    \label{eq:attended_feature}
\vspace{-0.25 em}
\end{equation}
where $\odot$ is the element-wise product between all channels in $F$ and the attention mask $P^{'}_{j}$. 
We concatenate $F^{'}_{j}$ for all $J$ parts and the background along the 3$^{rd}$ dimension as 
$F^{'} \in \mathbb{R}^{H\times W \times C^{*}} $, where $C^{*} = C^{'}  (J+1)$ and $C^{'} \neq C$, and pass it through a convolutional layer for per-pixel inference.

\qheading{Supervision}
We supervise the attention branch with part-segmentation maps (see ``dataset splits'' in \cref{sec:experiments_DataSplits} for the data source), and the contact branch with our contact annotations. 
Both branches classify pixels as being either a human part or the background; for the contact branch ``background" means ``no contact."
The overall loss is:
\vspace{-0.25 em}
\begin{equation}
    L = \lambda_a L_a + \lambda_c L_c   \text{,}
    \label{eq:loss}
\vspace{-0.25 em}
\end{equation}
where 
$L_a$ is a cross-entropy loss between the estimated attention maps and ground-truth part-segmentation maps, 
$L_c$ is a cross-entropy loss between the estimated and ground-truth contact maps, and 
$\lambda_a$ and $\lambda_c$ are steering weights.

\subsection{Implementation Details}
During training, body-part supervision for the attention branch is applied only in the initial stages, following Kocabas \etal~\cite{kocabas2021pare}; 
$\lambda_a$ is set to $0$ at later stages. 
We use a pre-trained dilated ResNet-50~\cite{yu2017dilated} as image encoder backbone. 
For the attention branch we use $3 \times 3$ convolutional layers with batch-norm and ReLU as image decoder. 
For the contact branch, we apply $3 \times 3$ convolutional layers with batch-norm and ReLU on the part-specific features. 
Since the background dominates the ground truth labels for both human part segmentation and contact estimation, we assign a smaller weight for the background label in the cross-entropy loss. 
For more details, see \supmat 

\section{Experiments}
\label{sec:experiments}

\begin{figure*}[t!]
	\centering
	\includegraphics[width=1.0\linewidth,trim={0cm 0cm 0cm 0cm},clip]{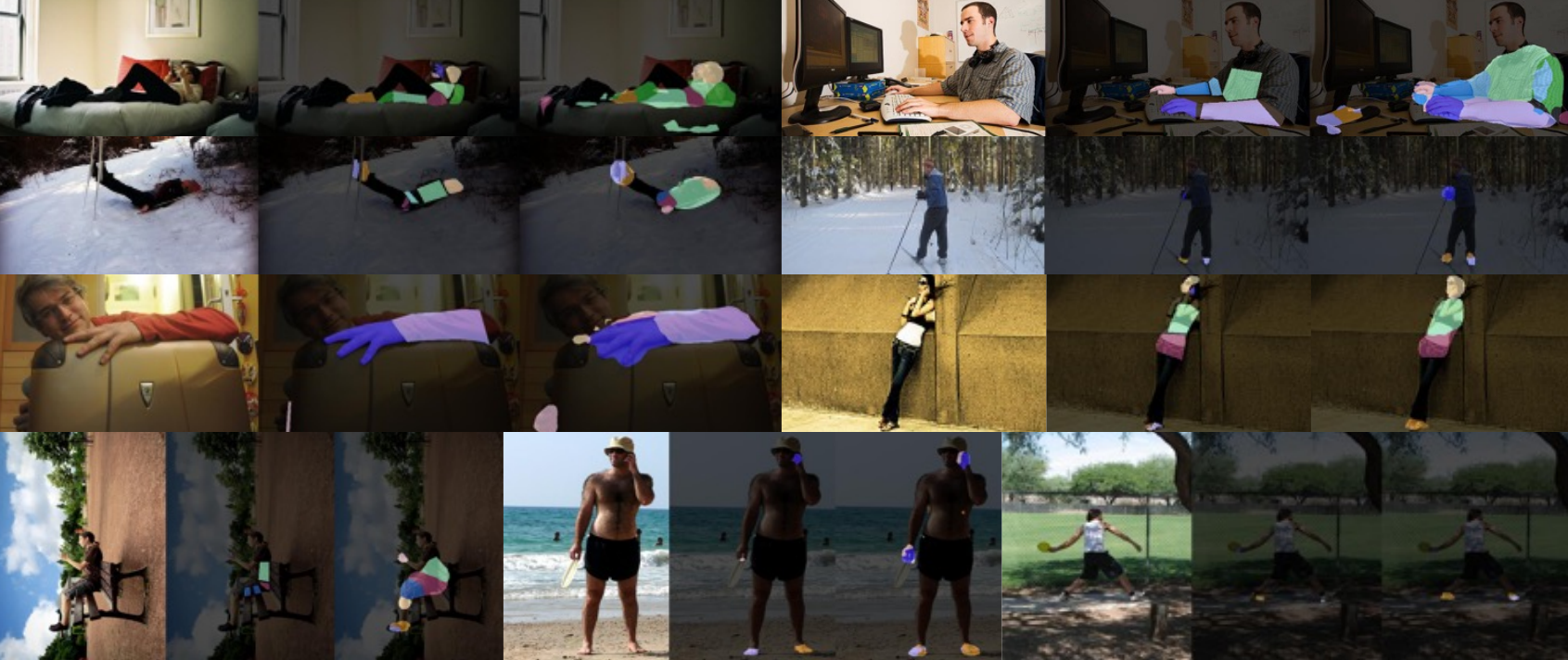}
        \vspace{-1.7em}
	\caption{
	            Qualitative results of our \hot contact detector for withheld images of our \hot dataset. 
	            For each input image we show the following triplet: 
	            \mbox{\tt\{Raw input image, Ground-truth contact, Predicted contact\}}. 
	}
	\label{fig:detector}
\end{figure*}

\subsection{Contact Detection} \label{sec:experiments_DataSplits}

\qheading{Dataset splits}
For \setAnnot, we randomly split the collected images into a training, validation and test set, resulting in $10,482$ images for training, $2,300$ for validation and $2,300$ images for testing. 
For \setGener, we split the training and testing set based on the scene; 
this results in $14,144$ images for training, $3,031$ for validation and $3,030$ for testing. 
To supervise the attention branch, we obtain pseudo ground truth for human part segmentation by rendering part-segmented \smplX meshes into per-part masks in image space; 
we use LEMO's~\cite{zhang2021lemo} \smplX fits for the \prox dataset \cite{hassan2019resolving} and use FrankMocap~\cite{rong2021frankmocap} to estimate \smplX meshes for the images of \setAnnot.

\qheading{Evaluation protocol}
We adopt the evaluation protocol of Zhou \etal~\cite{zhou2018semantic}; this is originally for semantic segmentation. 
We add one metric for contact prediction to evaluate whether the model distinguishes between contact and non-contact, \ie, ``background''.
We use the following metrics:
\begin{itemize}[leftmargin=*,noitemsep,nolistsep]
  \item[--] \emph{Semantic contact accuracy (SC-Acc.):} 
            The proportion of pixels that are both correctly classified as in contact and associated with the correct body-part label.
  \item[--] \emph{Contact accuracy (C-Acc.):} 
            The proportion of correctly classified pixels for binary contact labels; 
            this ignores the body-part label, in contrast to ``SC-Acc.'' 
  \item[--] \emph{Mean \acs{iou} (m\acs{iou})}: 
            The \ac{iou} between the predicted and the ground-truth contact pixels, averaged over all the body-part labels.
  \item[--] \emph{Weighted \acs{iou} (w\acs{iou})}: 
            m\acs{iou} weighted by the pixel ratio of each contact label. 
\end{itemize}

To study the influence of the \setAnnot and \setGener sets of the \hot dataset, we report performance by training and testing models separately on these, as well as on their combination that we denote as ``Full Set''. 
For the ``Full Set'', we randomly choose images from the \setGener so that the number of training and testing images from both sets are the same.

\qheading{Baselines}
There exists no model for full-body contact detection in images, other than ours. 
Thus, we evaluate contact detection for two models, ResNet$+$UperNet~\cite{xiao2018unified} and ResNet$+$PPM~\cite{zhao2017pyramid}, originally proposed for segmentation. 

\qheading{Ablations}
We evaluate two variants of our proposed model to ablate the contribution of the attention branch: 
``Ours$_{\text{wo/ att}}$''    without the attention branch and 
``Ours$_{\text{pure\_att}}$''  without supervision for the attention branch, which functions as an unsupervised pure soft-attention module.

\qheading{Results \& discussion}
Quantitative results for contact detection are shown in \cref{tab:contact_quantitative}, and qualitative results are shown in \cref{fig:detector}. 
Below we summarize key findings.

\begin{table*}
    \centering
    \resizebox{\hsize}{!}{
    \scriptsize
    \begin{tabular}{l|cccc|cccc|cccc}
        \toprule
        \multirow{2}{*}{\textbf{Model}} & 
        \multicolumn{4}{c|}{\shortstack[c]{\textbf{\setAnnot}}} & 
        \multicolumn{4}{c|}{\shortstack[c]{\textbf{\setGener}}} & 
        \multicolumn{4}{c}{\shortstack[c]{\textbf{``Full Set''}}} 
        \\
        \cline{2-13}
        &\textit{SC-Acc.}$\uparrow$  & \textit{C-Acc}$\uparrow$ &  \textit{mIoU}$\uparrow$ & \textit{wIoU}$\uparrow$ &\textit{SC-Acc.}$\uparrow$ & \textit{C-Acc}$\uparrow$  & \textit{mIoU}$\uparrow$ & \textit{wIoU}$\uparrow$& \textit{SC-Acc.}$\uparrow$  & \textit{C-Acc}$\uparrow$ & \textit{mIoU}$\uparrow$ & \textit{wIoU}$\uparrow$
        \\
        \midrule
        ResNet+UperNet~\cite{xiao2018unified}
        &35.1&62.6&0.195&0.227&21.1&42.7&0.080&0.116&32.5&62.4&0.187&0.214\\
        ResNet+PPM~\cite{zhao2017pyramid} &34.6&61.1&0.201&0.233&21.2&41.1&0.075&0.119&31.5&58.4&0.176&0.212\\
        Ours$_{\text{wo/ att}}$ &24.1&42.8&0.148&0.187&12.0&24.6&0.051&0.099&19.4&29.3&0.130&0.155\\
        Ours$_{\text{pure\_att}}$ &33.8&58.4&0.189&0.237&20.3&40.1&0.077&0.113&30.4&55.9&0.163&0.206\\
        Ours$_{\text{Full}}$ &\textbf{40.7}&\textbf{70.7}&\textbf{0.215}&\textbf{0.260}&\textbf{30.4}&\textbf{54.3}&\textbf{0.139}&\textbf{0.167}&\textbf{36.4} & \textbf{66.3 }&\textbf{0.209}& \textbf{0.251} \\
        \bottomrule
    \end{tabular}
    }
    \vspace{-0.5 em}
    \caption{Evaluation of contact detection accuracy on the \textsl{\hot} dataset.}
    \label{tab:contact_quantitative}
    \vspace{-0.8 em}
\end{table*}

\smallskip
\noindent 1.
Our model outperforms state-of-the-art (SOTA) methods \cite{xiao2018unified,zhao2017pyramid} developed for semantic segmentation (retrained for our task). 
This is due to the different nature of semantic scene understanding and contact estimation. 
The former relies on dense pixel annotations of the entire scene and global contextual features. 
The latter relies on sparser annotations and needs an attention mechanism to focus around humans. 

\smallskip
\noindent 2.
Our attention mechanism guides our model to learn better features that improve contact estimation. 
``$\text{Ours}_{pure\_att}$'',      which uses unsupervised pure soft-attention, outperforms 
``$\text{Ours}_{\text{wo/ att}}$''  which has no attention branch. 
By adding supervision on human-part segmentation in early training stages, the attention focuses on areas around each human part. 
Intuitively, this helps reasoning about contact by using both human-body and surrounding-object information; 
see \cref{fig:part_attention} for a visualization of the learned attention maps.

\smallskip
\noindent 3.
Learning on \setGener is more difficult than on \setAnnot.
This is partially because, even though we generate contact annotations from relatively ``clean'' \smplX fits by LEMO~\cite{zhang2021lemo}, which reasons about temporal continuity and occlusion, these are still a bit noisy. 
Fine-grained contact detection is sensitive to strong occlusions during interactions, motion blur, the low resolution for people observed by indoor-monitoring cameras, and the imperfect ``hallucination'' of SOTA pose estimation methods~\cite{rempe2021humor,zhang2021lemo} for resolving these ambiguities.
This shows the value of \setAnnot, \ie the collection of a high-quality dataset of \inthewild images with rich manual contact annotations, and points to important future work. 

\begin{table}[t]
\resizebox{1.0\linewidth}{!}{
\scriptsize
     \begin{tabular}{l l | c  c  c  c}
        \toprule
        \textbf{Train} & \textbf{Test}  & SC-Acc.$\uparrow$  & C-Acc.  $\uparrow$&  mIoU$\uparrow$&  wIoU$\uparrow$\\
        \midrule 
        HOT-Gen & HOT-Gen   & 30.4 &	54.3  &	0.139 & 0.167 \\
        HOT-Ann  & HOT-Gen   & 28.4 &	51.8  & 0.122 & 0.203 \\
        Full Set  & HOT-Gen   & \textbf{34.3} &	\textbf{59.2}  & \textbf{0.140} & \textbf{0.205} \\
        \midrule 
        HOT-Gen  & HOT-Ann   & 2.46 &	6.37  & 0.019 & 0.042 \\
        HOT-Ann  & HOT-Ann   & 40.7 &	70.7  &	0.215 & 0.260 \\
        Full Set & HOT-Ann & \textbf{47.4}&\textbf{79.2}&\textbf{0.232}&\textbf{0.273}\\
        \bottomrule 
     \end{tabular}
}
\vspace{-0.5 em}
\caption{Transfer across \setGener $\leftrightarrow$ \setAnnot.}
\vspace{-0.5 em}
\label{tab:supp_transfer}
\end{table}

\smallskip
\noindent 4. We conduct transfer experiments across \setAnnot and \setGener and the results are shown in \cref{tab:supp_transfer}. 
The model trained on \setAnnot generalizes well to \setGener, but not vice versa. 
This is mainly because the former is captured in the wild and has rich variation, while the latter is captured in constrained settings. 
It is also noteworthy that combining both datasets (``Full Set'') boosts performance, suggesting that automatic contact annotations like \setGener are beneficial for this task. 

We discuss the failure cases and report our model's performance under different settings, \ie, contact for different body parts and various contact area sizes; see \supmat

\subsection{Full-Body vs Part-Specific Contact} \label{sec:experiments_PartSpecific}
\begin{figure}
\centering
\begin{subfigure}[]{\linewidth}
    \includegraphics[width=0.495\linewidth,trim={0cm 0cm 0cm 0cm},clip]{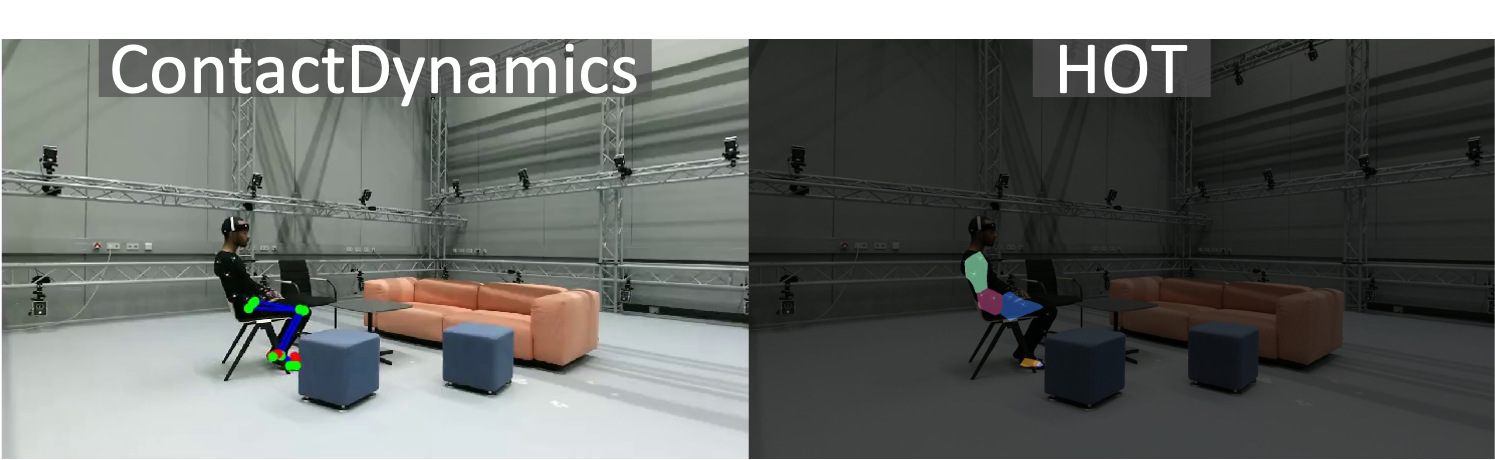}
	\includegraphics[width=0.495\linewidth,trim={0cm 0cm 0cm 0.58cm},clip]{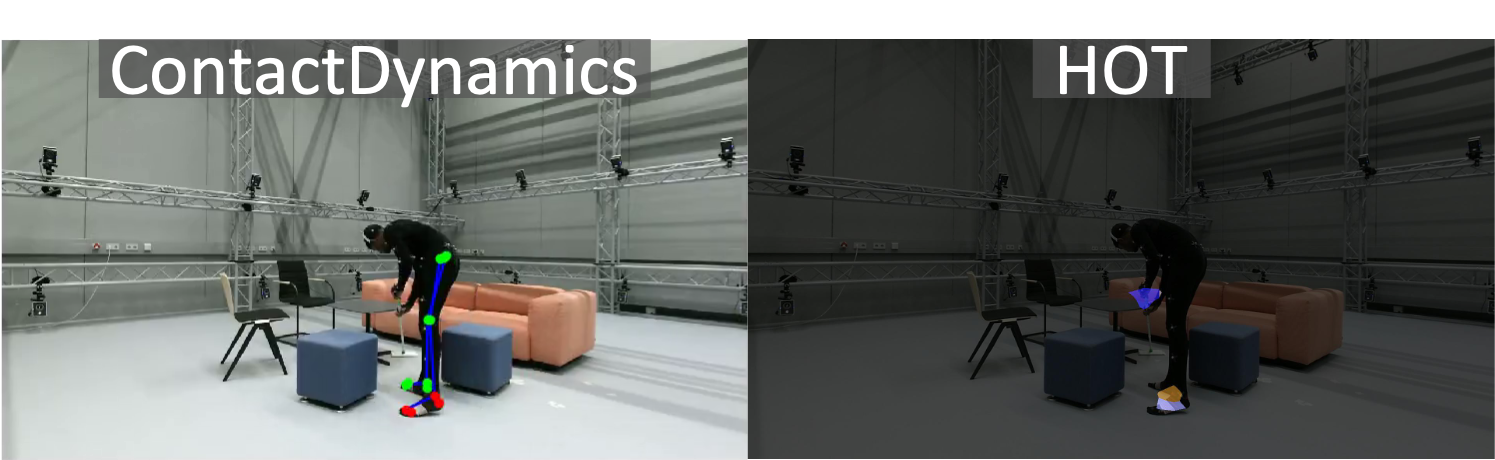}
\end{subfigure}
\\
\begin{subfigure}[]{\linewidth}
	\includegraphics[width=0.495\linewidth,trim={0cm 0cm 0cm 0cm},clip]{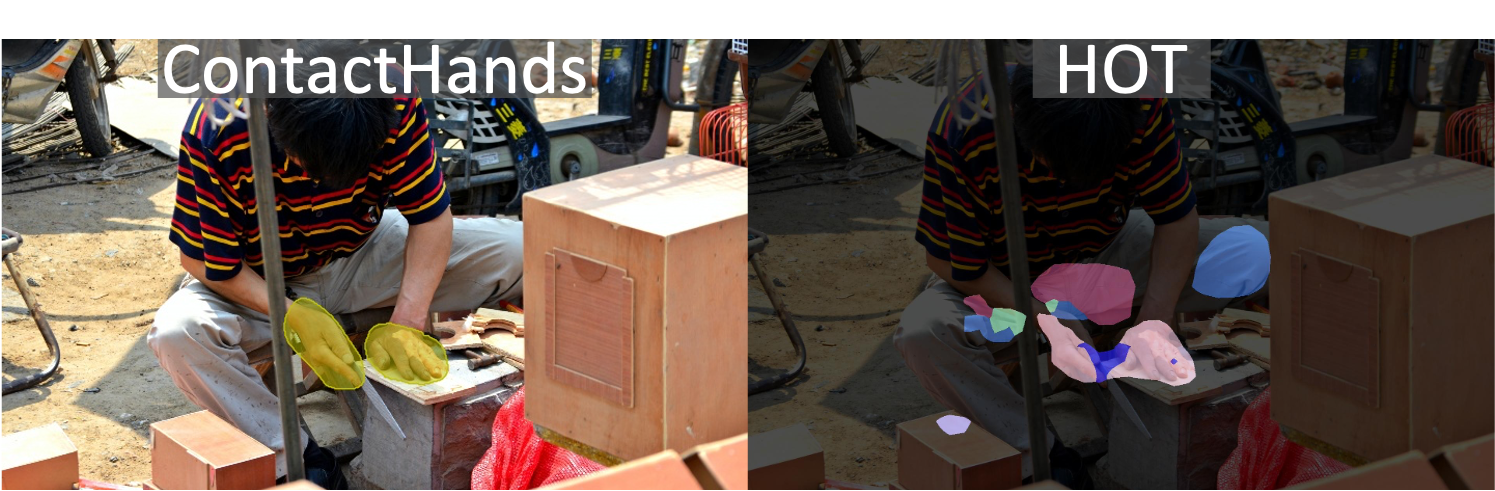}
    \includegraphics[width=0.495\linewidth,trim={0cm 0cm 0cm 0cm},clip]{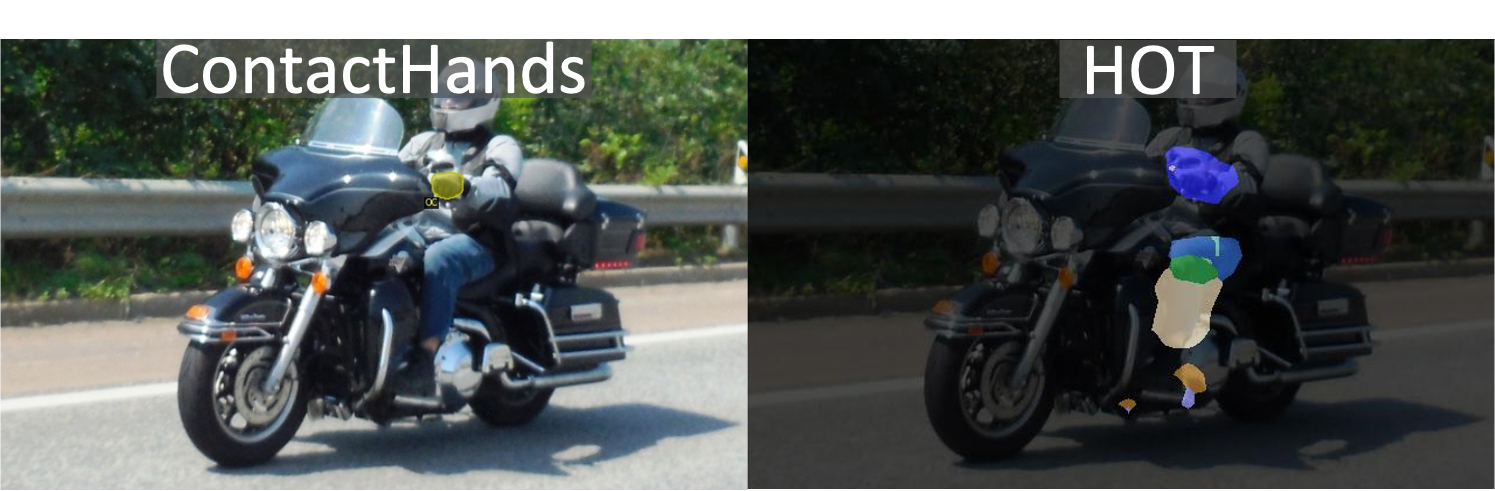}
\end{subfigure}
\vspace{-0.7 em}
\caption{
                Comparison of our general-purpose full-body contact detector (\hot) against existing part-specific detectors, namely
                ``ContactDynamics''~\cite{rempe2020contact} for joint-level foot-ground contact, and 
                ``\mbox{ContactHands}''~\cite{narasimhaswamy2020detecting} for bounding-box-level hand contact. 
}
\label{fig:part_contact}
\vspace{-0.2 em}
\end{figure}

To evaluate the robustness of our general-purpose full-body \hot contact detector, we compare it against two existing part-specific contact detectors, as shown in \cref{fig:part_contact}:

\qheading{(i) Foot contact}
``ContactDynamics''~\cite{rempe2020contact} estimates \emph{joint-level} foot-ground contact from a \emph{video}.
We evaluate our model and ``ContactDynamics'' against the ground-truth foot contact from \prox's ``quantitative set''. 
Our detector achieves a similar performance (\hot $\textbf{59.2}\%$ vs ``ContactDynamics'' $58.6\%$). Thus, 
it could be a drop-in replacement contact detector for \threeD body pose estimation \cite{rempe2020contact}.

\qheading{(ii) Hand contact}
``ContactHands''~\cite{narasimhaswamy2020detecting} detects hands and classifies them into 
``self-contact'', 
``person-person contact'', and 
``person-object contact''. 
We evaluate our \hot detector and ``ContactHands'' on hand-object contact on a subset of the \setAnnot test set. 
We report contact recognition accuracy under an \ac{iou} threshold of 0.4; 
our detector achieves similar performance (\hot $\textbf{63.5}\%$ vs ~\cite{narasimhaswamy2020detecting} $62.2\%$). 

\begin{table}
\centering
\scriptsize

     \begin{tabular}{l | c  c  c  c  c}
        \toprule
        ~\textbf{Method}~   & ~No Contact~ & \prox~\cite{hassan2019resolving} &  ~Predicted~Contact~ &  ~GT~Contact~ \\
        \midrule 
        V2V $\downarrow$    & 183.3 &	174.0  &	                  \textbf{172.3} & 163.0 \\
        \bottomrule 
     \end{tabular}
    \vspace{-1.0 em}
\caption{Contact-driven human pose estimation performance.}
\label{tab:contact_hps}
\vspace{-0.5 em}
\end{table}

\qheading{Discussion}
As shown in \cref{fig:part_contact}, ``\mbox{ContactDynamics}'' simply classifies foot joints as in contact or not, and ``\mbox{ContactHands}'' detects hands as bounding boxes, while we generalize to the full body and detect richer heatmaps.
In \supmat, we provide more details and some preliminary results when testing our model on self-contact and human-human contact.
The fact that our full-body contact detector performs on par with existing part-expert detectors
holds promise for developing a general purpose contact detector for human-object and human-human interactions in future work.

\begin{table}
\centering
\scriptsize
     \begin{tabular}{l l | c  c  c}
        \toprule
        \textbf{Train} & \textbf{Test}  & precision  $\uparrow$  & recall  $\uparrow$&  f1 $\uparrow$\\
        \midrule 
        RICH~\cite{huang2022capturing}  & RICH   & \textbf{0.699} &	0.744  &	                  \textbf{0.708} \\
        RICH+HOT-pGT  & RICH   & 0.675 &	\textbf{0.761}  &	                  0.684 \\
        \midrule 
        RICH  & HOT   & 0.439 &	0.192  &	                        0.231 \\
        RICH+HOT-pGT  & HOT   & \textbf{0.684} &	\textbf{0.701}  &	                  \textbf{0.636} \\
        \bottomrule 
     \end{tabular}
\vspace{-0.5em}
\caption{\threeD dense contact estimation performance.}
\label{tab:contact_3D}
\vspace{-1.0em}
\end{table}

\subsection{\hot Contact Detection vs  Heuristic Contact}
\label{sec:experiments_PROX_like}

The \prox dataset \cite{hassan2019resolving} is widely used for developing and evaluating \ac{hoi} methods. 
Its human meshes have been reconstructed with an optimization method that fits \smplX to images, paired with an a-priori known (\ie, pre-scanned) \threeD scene.
The human meshes look physically plausible, as the method encourages (manually annotated) ``likely contact'' body vertices that lie close to the scene to contact it while not penetrating it; this resolves pose errors. 

We replace \prox's manually annotated ``likely contact'' vertices with the ones of the body parts that our detector suggests are in contact, given the input image. 
We call this setup ``Predicted Contact'' and evaluate this on \prox's ``quantitative set'' via the Vertex-to-Vertex (\emph{V2V}) error. 
We also evaluate a baseline with \mbox{``No Contact''} constraints. 
For a fair comparison, for all baselines we use the same optimization process as in \prox~\cite{hassan2019resolving}. 
Results in \cref{tab:contact_hps} show that our ``Predicted Contact'' is on par with ``\prox'', indicating that detecting contact in images is promising for replacing \prox's handcrafted heuristics.
We also simulate a perfect contact detector using \prox's ground truth (``GT Contact''). 
This shows that there is room and merit for improving image-based contact detection in future work. 

\subsection{\hot for \threeD Contact on Bodies}
\label{sec:experiments_BSTRO_like}

\begin{figure}
	\centering
	\includegraphics[width=1.0\linewidth,trim={0cm 0cm 0cm 0cm},clip]{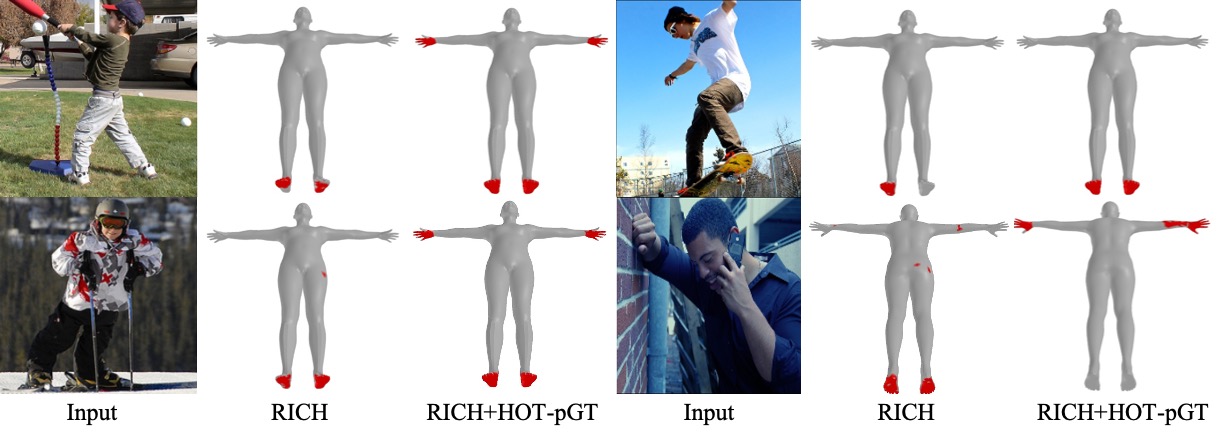}
    \vspace{-2.0em}
	\caption{
	            Qualitative results of 3D contact estimation on \hot.
	}
    \vspace{-0.9 em}
	\label{fig:contact3d}
\end{figure}

The recent RICH dataset and BSTRO model \cite{huang2022capturing} focus on dense \threeD contact estimation on the human body from an image.
To show the usefulness of our \hot dataset for this task, we ``lift'' our 2D annotations to coarse 3D ones, by annotating the respective 3D \smpl parts as in contact, and treat these as pseudo ground-truth (\mbox{\hot-pGT}). 
We then employ the BSTRO model \cite{huang2022capturing}, extend its training dataset with \mbox{\hot-pGT}, and re-train. 
We report performance in \cref{tab:contact_3D} and \cref{fig:contact3d}. 
The model trained on RICH data alone fails on \hot images, which are taken in the wild, in contrast to RICH images.
Interestingly, adding \mbox{\hot-pGT} for training improves the \threeD contact estimation accuracy for \inthewild images, while not hurting for RICH data.
This shows that our \hot dataset can help for \threeD contact estimation and related applications.
For details, see \mbox{\supmat}

\section{Conclusion}
\label{sec:conclusion}

We focus on human-object contact detection for images. 
To this end, we collect the \hot dataset and develop the \hot contact detector with human-part guided attention. 
Our detector outperforms baseline models and generalizes reasonably well for in-the-wild images. 
One limitation is that we build our method upon ``simple'' convolutional models; however our key insight (human part attention) is general and agnostic to the model's architecture. 
We believe that this new task, dataset and model fill a gap in the literature and hope they can inspire more future endeavors into this topic, utilizing more complex models like transformers~\cite{cheng2021per,cheng2022masked} and exploring a wider range of applications.

{\small
\qheading{Acknowledgments}
We thank Chun-Hao Paul Huang for his valuable help with the RICH data and the training code of BSTRO \cite{huang2022capturing}. 
We thank Lea M\"{u}ller, Mohamed Hassan, Muhammed Kocabas, Shashank Tripathi, and Hongwei Yi for the insightful discussions. 
We thank Benjamin Pellkofer for IT help, and Nicole Overbaugh and Johanna Werminghausen for administrative help. 
This work was partially funded by the German Federal Ministry of Education and Research (BMBF): T{\"u}bingen AI Center, FKZ: 01IS18039B.
DT's work was mostly performed at the MPI-IS.
}

{
    \small
    \qheading{Disclosure}
    \href{https://files.is.tue.mpg.de/black/CoI_CVPR_2023.txt}{
         https://files.is.tue.mpg.de/black/{CoI\_CVPR\_2023.txt}}
}

{\small
\balance
\bibliographystyle{config/ieee_fullname}
\bibliography{config/BIB}
}

\clearpage
\appendix

\renewcommand{\thefigure}{A.\arabic{figure}}
\renewcommand{\thetable}{R.\arabic{table}}
\renewcommand{\theequation}{A.\arabic{equation}}
\setcounter{figure}{0}
\setcounter{table}{0}
\setcounter{equation}{0}

\section*{Appendix}

In \cref{sec:supp_human_part},   we introduce the detailed human-body part labels for human-object contact.
In \cref{sec:supp_dataset}, we describe more details for the annotation protocol for \setAnnot and how we generate pseudo ground truth for \setGener. 
In \cref{sec:supp_implement}, we report more implementation details. 
\cref{sec:supp_contact_detect} shows more experimental results in the contact detection task, including failure cases, evaluation under different settings and attention maps, \etc.
In \cref{sec:supp_part_specific_contact}, we provide more details of the part-specific contact detector that we compare with \hot.
In \cref{sec:supp_hps_results}, we report more experiment details and results to illustrate the use of our \hot contact detection for \threeD human pose estimation. 
\cref{sec:supp_3d_contact} includes more details on how the \hot dataset can facilitate 3D contact estimation.
\Cref{sec:supp_downstream_application} discusses more potential downstream applications for contact detection and qualitative results on self-contact and human-human contact.
The use of existing assets is listed in \cref{sec:supp_assets}.

\section{Human Part Labels}
\label{sec:supp_human_part}

For the contact estimation task, we want to know if contact takes place in the image, the area in which it takes place, as well as the body part that is involved.

\begin{figure}[ht!]
	\centering
	\includegraphics[width=0.9\linewidth,trim={0cm 0cm 0cm 0cm},clip]{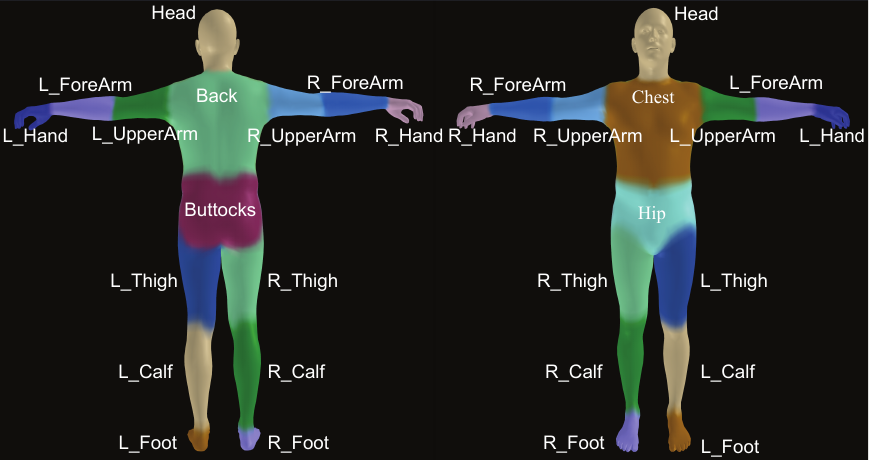}
	\caption{
	        The color-coded human parts with labels.
	}
	\label{fig:supp_human_part}
\end{figure}

To get the human part labels, we divide the parametric human body model, \smplX~\cite{smplifyPP} into $17$ parts, 
\ie:  Head, Chest, L$\_$UpperArm, L$\_$ForeArm, L$\_$Hand, R$\_$UpperArm, R$\_$ForeArm, R$\_$Hand, Buttocks, Hip, Back,  L$\_$Thigh, L$\_$Calf, L$\_$Foot, R$\_$Thigh, R$\_$Calf and R$\_$Foot.
This is based on the original part segmentation of \smplX, but for simplicity we unite certain parts (\eg, parts of the back across the spine), that even human annotators cannot easily differentiate.
\Cref{fig:supp_human_part} shows the color-coded body parts, together with part labels, on the \smplX mesh.

\section{Dataset Details}
\label{sec:supp_dataset}
\subsection{Contact Annotation for \setAnnot}
\label{sec:supp_contact_anno}

We hire professional annotators to annotate the contact information for the in-the-wild images. 
The annotation pipeline is similar to semantic segmentation annotation but with different task requirements. 
In this section, we describe the instructions given to the annotators in detail. 

The overall annotation process includes two steps: 
(1)     ``segmenting'' the image area for human-object contacts, and 
(2)     assigning the human part label associated with the contact. 
In the first step, the annotators are asked to hallucinate the contact area in an image and draw a tight polygon around it. 
In the second step, the annotators pick a label for the contact area out of our pre-defined $17$ human parts.

Determining the exact contact area between a human and an object is non-trivial, especially in the image space. 
Thus, we first perform a round of trial annotations, in which we test our annotation protocol, as well as train our annotators. 
We provide the following instructions to annotators:
\begin{itemize}[nolistsep] 
    \item[--]   Contact areas between humans and objects are always occluded. 
                Annotators should hallucinate the contact area in \threeD, and then annotate its projection on the \twoD image. 
    \item[--]   A polygon annotation should cover only the subset of the human part that is in contact, 
                and not the whole part. 
                Note that this is different from part segmentation.
    \item[--]   There may be multiple contact areas between a single human and a single object. 
    \item[--]   Only humans in the foreground should be considered; any humans in the background should be ignored. 
    \item[--]   Contact areas that are occluded by another human or object should be ignored. 
    \item[--]   Contact for body parts with extreme out-of-frame cropping, \eg, when only a hand is visible, 
                should be ignored. 
    \item[--]   Human-human and self contact should be ignored.
\end{itemize}
After a full annotation round, we have two rounds of quality checks. 
In more detail, for every $3$ annotators, there is $1$ extra annotator that only conducts quality checks. 
The quality check verifies if the annotated polygon matches the contact area, if the contact label corresponds to the correct body part, if there are missing contact annotations (false negatives), if there are false positive contact annotations and if contact annotations are consistent across images. 
\begin{figure}
	\centering
	\includegraphics[width=1.0\linewidth,trim={0mm 10mm 0mm 09mm},clip]{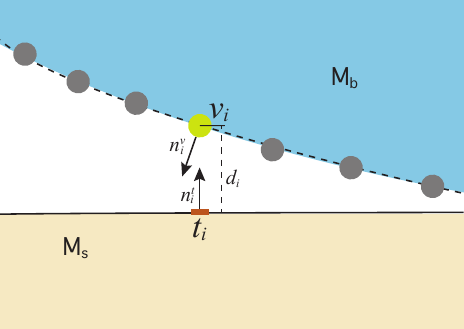}
	\caption{
	      Illustration of computing the properties involved in the contact annotation between the body mesh $\mathcal{M}_b$ and scene mesh $\mathcal{M}_s$ for \setGener. 
	}
	\vspace{-0.8 em}
	\label{fig:supp_contact}
\end{figure}

\subsection{Contact Generation for \setGener}
\label{sec:supp_contact_gen}

The \prox dataset \cite{hassan2019resolving} captures human subjects interacting with static scenes. 
Briefly, we use the reconstructed \threeD human and scene meshes to first compute the human vertices that are in close \threeD proximity to scene ones, and consider the former as contact vertices. 
We then render the respective triangles onto the \twoD image to get automatic contact area annotations, as well as the associated body labels. 

More specifically, the human pose and shape is represented with the \smplX body model with pose parameters, $\theta$, and shape parameters, $\beta$. 
The \threeD human mesh is denoted as $\mathcal{M}_b \in \mathbb{R}^{10475\times3}$. 
Each vertex, $v_i \in \mathbb{R}^{3}$, has a surface normal $n_{i}^{v}$ and an associated human part label $c_i$. 
For each frame, given the estimated \smplX mesh, $\mathcal{M}_b$, and the scene mesh, $\mathcal{M}_s$, we first calculate the distance $\{d_i\}_{i=1}^{10475}$ from all human vertices $\{v_i\}_{i=1}^{10475}$ to the scene mesh $\mathcal{M}_s$. 
For each vertex $v_i$, we also find the closest triangle in $\mathcal{M}_s$, denoted as $t_i$, with surface normal $n_i^t$.

Then, a human vertex, $v_i$, is considered in contact if its distance to the scene, $d_i$, is below a threshold, 
and the surface normal, $n_i^v$, is in the opposite direction to the scene normal, $n_i^t$. 
Specifically, both of the following two constraints should be satisfied:
\begin{itemize}[noitemsep,nolistsep]
    \item[--] \textbf{Distance constraint:} $d_i \leq \delta_d$, where the distance threshold $\delta_d$ is set to be $0.07 m$ empirically;
    \item[--] \textbf{Surface normal compatibility:} $\texttt{Angle}(n_i^v, n_i^t) \geq \delta_a$, where the $\delta_a = 110^{\circ}$ is an angle threshold. 
\end{itemize}
\Cref{fig:supp_contact} demonstrates the criteria mentioned above.

Finally, for the contact vertices we find the respective triangles on the \threeD body mesh, and render them separately per body part to get dense \twoD contact areas. 
In this way, we automatically create pseudo ground truth for contact. 

\subsection{Annotation repeatability in \setAnnot}
Annotating contact from images is a very challenging task. To verify the  repeatability of the manual annotation, two new trained persons are hired to
annotate 200 random images from \setAnnot. 
We compare the labels to the ones collected by the annotators of \cref{sec:supp_contact_anno}. 
The agreement for body-part contact labels is $93.2\%$, and 
the agreement for pixel contact labels is $77.1\%$; this is comparable to the $82.4\%$ agreement of the semantic-segmentation pixel annotations of ADE20K~\cite{zhou2018semantic} in their experiment for consistency check across annotators.

\subsection{Dataset Statistics by Splits}
\begin{figure}[ht!]
\vspace{-3mm}
\begin{subfigure}[t]{\linewidth}
\centering
\includegraphics[width=0.9\linewidth,trim={0cm 0cm 0cm 0cm},clip]{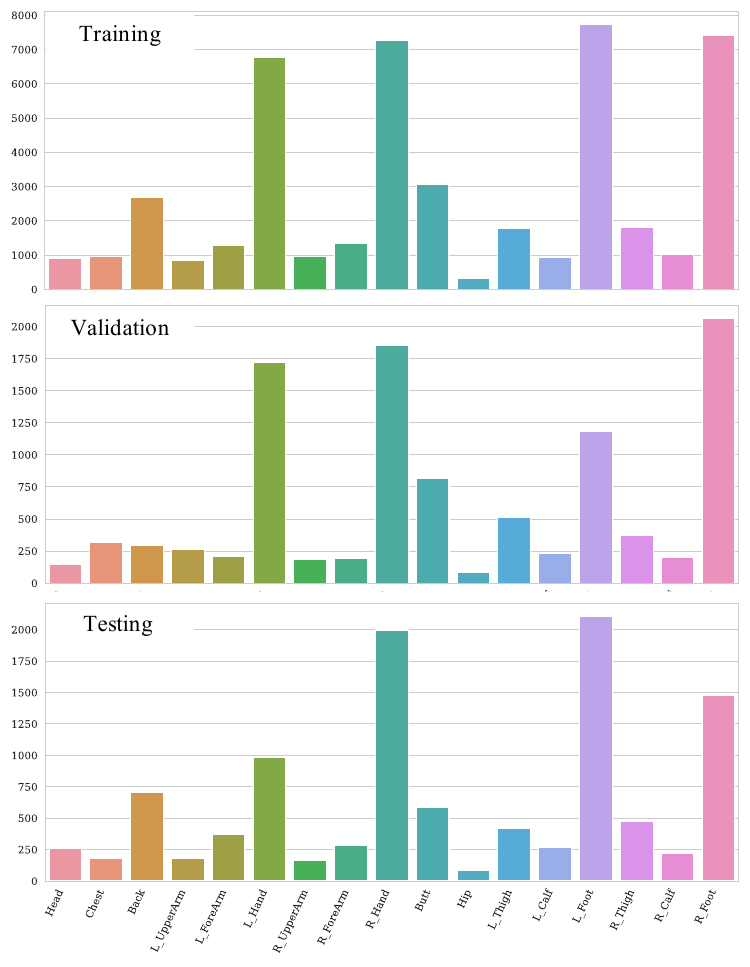}
\end{subfigure}%
\caption{Distribution of body-part labels for contact in \setAnnot; number of contact areas (Y-axis) for a certain body part (X-axis) in different data splits.}
\label{fig:supp_hot_split_stats}
\end{figure}

\begin{figure}[ht!]
\vspace{-3mm}
\begin{subfigure}[t]{\linewidth}
\centering
\includegraphics[width=0.9\linewidth,trim={0cm 0cm 0cm 0cm},clip]{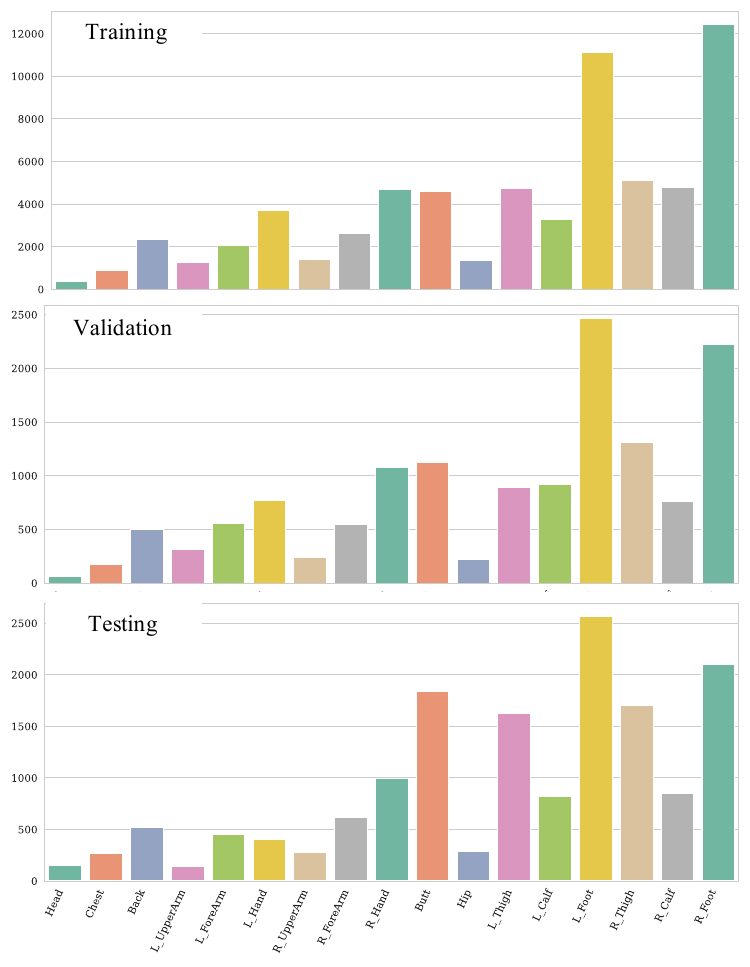}
\end{subfigure}%
\caption{Distribution of body-part labels for contact in \setGener; number of contact areas (Y-axis) for a certain body part (X-axis) in different data splits.}
\label{fig:supp_prox_split_stats}
\end{figure}

Current \ac{hoi} datasets have  many walking, standing-up, or sitting-down poses (foot contact) or grasping poses (hand contact); 
this naturally biases the data distributions as shown in the main paper. 
Randomly spliting data into training, validation and testing sets naturally captures such biases, but the statistics are similar across these sets as can be seen from \cref{fig:supp_hot_split_stats,fig:supp_prox_split_stats}.

\section{Implementation Details}
\label{sec:supp_implement}

During training, the loss weight for the attention branch $\lambda_a$ is set to be $0.1$ for the first $10$ epochs and $0$ for the rest of the epochs. 
The loss weight $\lambda_c$ for contact estimation is set to be $1$. 
We use a pre-trained dilated ResNet-50~\cite{yu2017dilated} as image encoder backbone. 
For the attention branch we use $3 \times 3$ convolutional layers with batch-norm and ReLU as image decoder, followed by another convolutional layer with kernel size 1 to make pixel-wise human part label classification. 
For the contact branch, we apply $3 \times 3$ convolutional layers with batch-norm and ReLU on the part-specific features, which we further concatenate along the channel axis. 
The weights of convolutional layers are different across human parts, so that the contact branch learns part-specific features under the attention guidance. 
Another convolutional layer with kernel size 1 is used to make pixel-wise contact label prediction. 
Since the background dominates the label ground truth for both human-part segmentation and contact estimation, we assign a smaller weight $0.02$ for the background label and $1$ for the rest of the labels in the cross-entropy loss. 
We re-scale all images to have their longer side $400$ pixels long, and then pad, if necessary. 
Random flipping is applied for data augmentation. 
We train the model for $20$ epochs on $4$ NVIDIA-A100 GPUs with a batch size of $24$. 
We use the SGD~\cite{robbins1951stochastic} optimizer, with an initial learning rate of $0.02$ with polynomial decay following Zhou \etal~\cite{zhou2018semantic}. 

We also report the model size for fair performance comparison during the experiments. 
Our model has a total of 50.2 million trainable parameters, whereas ResNet$+$PPM~\cite{zhao2017pyramid} has 46.7 million  and ResNet$+$UperNet~\cite{xiao2018unified} has  64.2 million. 

\section{More Contact Detection Results }
\label{sec:supp_contact_detect}

\subsection{Failure Cases}
\label{sec:supp_fail}

\begin{figure}[ht!]
	\includegraphics[width=1.0\linewidth,trim={0cm 0cm 0cm 0cm},clip]{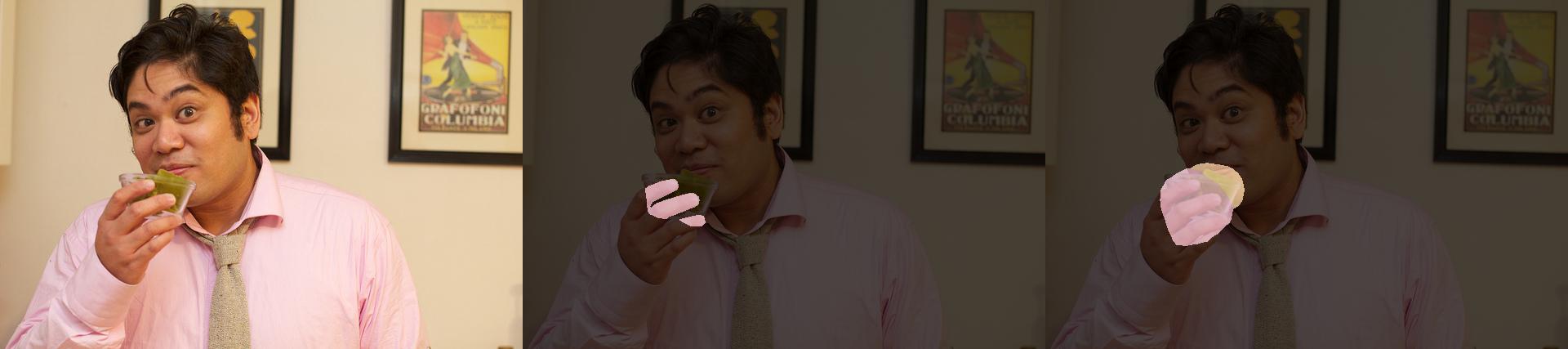}
	\\
	\includegraphics[width=1.0\linewidth,trim={0cm 0cm 0cm 0cm},clip]{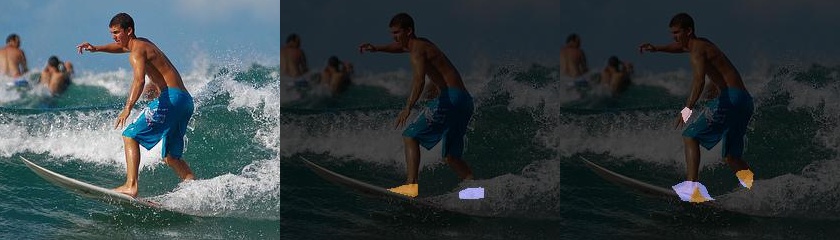}
	\\
    \includegraphics[width=1.0\linewidth,trim={0cm 0cm 0cm 0cm},clip]{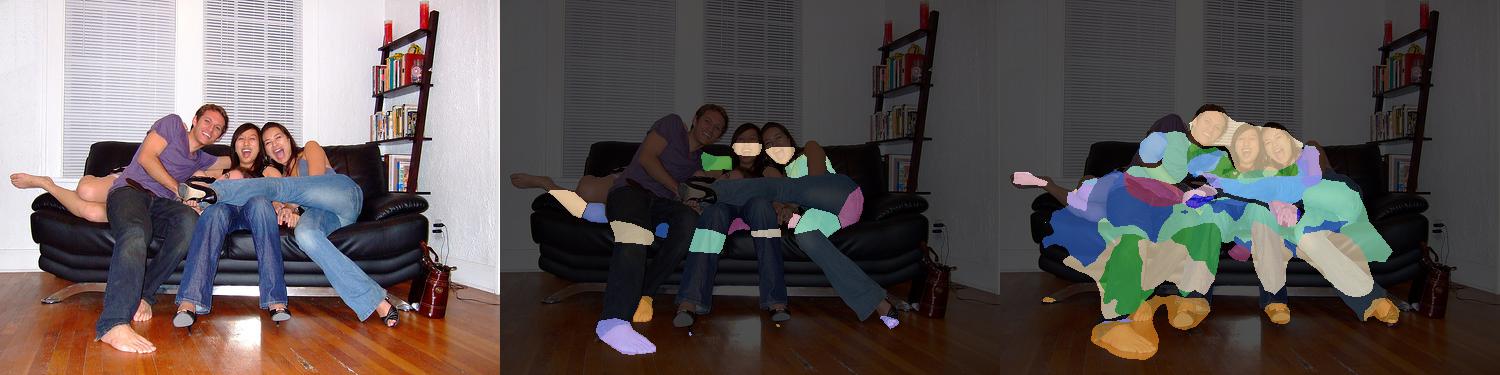}
	\vspace{-0.5 em}
	{\small \hspace*{0.9cm}  Input \hspace{2.15cm}   GT \hspace{1.6cm} Prediction \hfill}\\
	\vspace{-1.5 em}
	\caption{
	            Representative failure cases for our contact detector.
	}
	\label{fig:fail_qual}
\vspace{+0.5 em}
\end{figure}

\Cref{fig:fail_qual} shows some examples of failure cases. 
We see that our model might struggle with occlusions, multiple persons or fine-grained contact areas.
We also observe that the model sometimes fails in distinguishing left and right for the body parts. 
These point out that contact detection may benefit from future work on adding human pose information, multi-resolution reasoning and differentiating human-object contact with self-contact and person-person contact, but these are currently out of our scope.

\begin{figure*}[t]
\includegraphics[width=\linewidth,trim={0cm 0cm 0cm 0cm},clip]{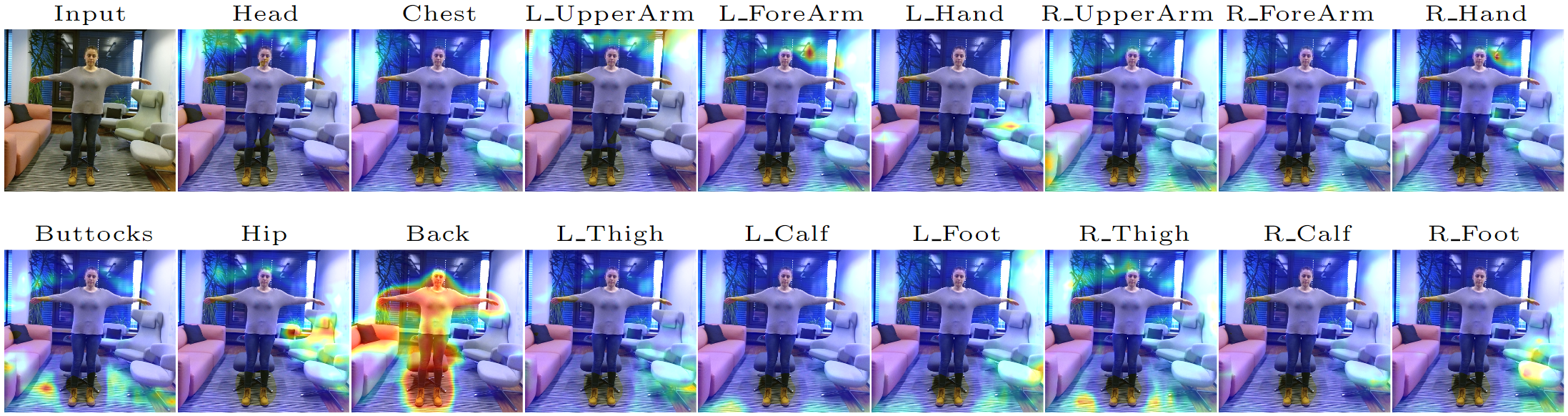}
\vspace{-1.5em}
\caption{Attention maps for ``Ours$_{\text{pure\_att}}$'', visualized separately per body part.}
\label{fig:supp_pure_attention}
\vspace{-0.5em}
\end{figure*}

\begin{table*}[ht]
\resizebox{1.0\linewidth}{!}{
\begin{tabular}{l|ccccccccccccccccc|c}
\toprule
body part & Head  & Chest & Back  & L\_UpperArm & L\_ForeArm & L\_Hand & R\_UpperArm & R\_ForeArm & R\_Hand & Butt  & Hip   & L\_Thigh & L\_Calf & L\_Foot & R\_Thigh & R\_Calf & R\_Foot & Mean  \\
\midrule
SC-Acc. $\uparrow$   & 54.9  & 27.4  & 62.0  & 29.3        & 11.4       & 43.1    & 5.07        & 2.86       & 69.5    & 57.0  & 3.77  & 12.0     & 20.3    & 47.5    & 11.3     & 7.95    & 36.4    & 40.7  \\
mIoU  $\uparrow$    & 0.532 & 0.252 & 0.558 & 0.199       & 0.092      & 0.215   & 0.047       & 0.026      & 0.430   & 0.374 & 0.034 & 0.173    & 0.138   & 0.334   & 0.090    & 0.070   & 0.262   & 0.260 \\
\bottomrule
\end{tabular}
}
\caption{Contact estimation performance by different body parts on \setAnnot.}
\label{tab:supp_contact_bypart}
\end{table*}
\subsection{Model Performance under Various Settings}
To better diagnose the model's performance under different settings, we conduct the following two experiments.

\smallskip
\noindent1) The contact detection for different body parts. Quantitative results are shown in \cref{tab:supp_contact_bypart}. We can see that our methods performs better on the body parts with more data, \eg, hand, foot and butt, and fails in the body parts that naturally have less contact, \eg, hip and calf. This shows the importance of data balance when developing a general purpose contact detector.

\noindent2) We also evaluate the model's performance with various contact area sizes, \ie, \textit{small}, \textit{medium} and \textit{large}. The size thresholds are $0.052\%$ and $0.22\%$ based on the size distribution, which can be seen in the main paper. The quantitative results in \cref{tab:supp_contact_bysize} show our model has decent performance on contacts with \textit{medium} and \textit{large} sizes, but cannot distinguish fine-grained contact with \textit{small} areas. This indicates that contact detection will benefit from multi-resolution reasoning for different types of human-object contact.

\begin{table}[ht]
\centering
\resizebox{0.75\linewidth}{!}{    
\scriptsize

     \begin{tabular}{l |  c  c  c}

        \toprule
        Contact area & Sc-Acc.$\uparrow$ & mIoU$\uparrow$  & wIoU$\uparrow$  \\
        \midrule
small        & 21.6    & 0.020 & 0.025 \\
medium       & 39.7    & 0.253 & 0.301 \\
large        & 53.4    & 0.381 & 0.494 \\
        \midrule
all          & 40.7    & 0.215 & 0.260\\
        \bottomrule 
     \end{tabular}
}
\caption{Contact estimation performance by contact area sizes on \setAnnot.}
\label{tab:supp_contact_bysize}
\end{table}

\subsection{Attention without Human Part Supervision}
\label{sec:supp_attention_compare}
\Cref{fig:supp_pure_attention} shows the learned attention maps for ``Ours$_{\text{pure\_att}}$''. 
In this setting, no supervision is applied for the attention branch, which functions as an unsupervised pure soft-attention module. 
In contrast to ``Ours$_{\text{Full}}$'' where the attention focuses on areas around each human part (see \colorRef{Fig.~5} in the main paper), for ``Ours$_{\text{pure\_att}}$'' certain parts (\eg, the ``Back'' in this case) attend to the full body, while others can get distracted by the background.

\section{Part-Specific Contact Detectors}
\label{sec:supp_part_specific_contact}

\subsection{Foot-Contact Detector}

``ContactDynamics"~\cite{rempe2020contact} is a physics-based trajectory optimization method that generates physically-plausible motions. 
To this end, an intermediate step detects contact for the \emph{toe} and \emph{heel} joints of each foot. 
The authors use \mocap sequences to generate ground-truth contact for training such a detector using heuristics. The contact detector is a multi-layer perceptron (MLP) that takes as input lower-body 2D joints in a temporal window, and outputs four contact labels (left/right toe, left/right heel) for the central frames.

For evaluation on \prox's test set (aka ``quantitative set''), we use OpenPose~\cite{cao2019openpose} to generate 2D keypoints and feed these into the pre-trained foot contact model. 
For a fair comparison with our \hot contact detector, we consider a foot to be in contact when at least one joint (either \emph{toe} or \emph{heel}) is in contact. Our detector achieves similar performance (\hot $\textbf{59.2}\%$ vs ContactDynamics~\cite{rempe2020contact} $58.6\%$); 
see the related discussion in \colorRef{\mbox{Sec. 5.2 (i)}} of the main paper. 
\begin{figure*}
\begin{subfigure}[t!]{\linewidth}
		\includegraphics[width=\linewidth,trim={0cm 0cm 0cm 0cm},clip]{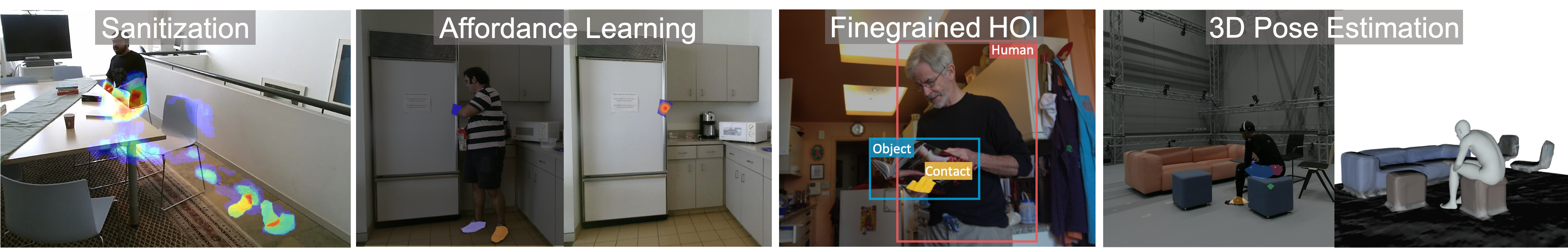}
\end{subfigure}
\caption{Example downstream applications of contact detection.}
\label{fig:downstream}
\end{figure*}

\subsection{Hand-Contact Detector}

``ContactHands''~\cite{narasimhaswamy2020detecting} detects hands as bounding boxes and classifies their contact state as ``self-contact'', ``person-person'', or ``person-object'' (hand-object)  contact. 
Here we only consider the hands with hand-object contact label in the model output. 

During evaluation, a detected hand-object contact from ``ContactHands'' is considered as a true positive if the hand bounding box and the ground-truth hand contact area overlap. 
For \hot, we consider our predicted hand contact area as a true positive if the \ac{iou} with the ground-truth hand contact area is larger than 0.4.
Experimental results show that our detector achieves similar performance (\hot $\textbf{63.5}\%$ vs ContactHands~\cite{narasimhaswamy2020detecting} $62.2\%$); 
see the related discussion in \colorRef{\mbox{Sec. 5.2 (ii)}} of the main paper. 

\section{\texorpdfstring{\hot}{} for 3D HPS Estimation}
\label{sec:supp_hps_results}
In the main paper, we replace the heuristic contact in \prox~\cite{hassan2019resolving} with our contact detection when estimating \threeD humans from a color image.
This tests the usefulness of our contact estimates for human pose estimation.
In \cref{tab:supp_contact_hps} we report the full-performance comparison on PROX's ``quantitative set'';
\textit{``All Contact''} considers all body vertices to be in contact.

Importantly, note that \emph{V2V} is the most appropriate ``pose'' metric for \emph{surface contact}, as vertices lie \emph{on surfaces} that come in contact with objects.
V2V numbers in \cref{tab:supp_contact_hps} show that detecting contact in images is promising and can be used to replace PROX's hand-crafted contact heuristics. 

The rest of the metrics do \emph{not} capture contact; they are reported for completeness. 
Procrustes (Pr.) \emph{factors out global translation and rotation} to focus only on articulation; 
``pr.PJE'' and ``pr.V2V'' are irrelevant for contact. 
Skeleton joints (PJE) lie \emph{under the surface} of the body.

\begin{table}[bh]
\centering
\smallskip
\resizebox{\linewidth}{!}{
\smallskip
     \begin{tabular}{l | c  c  c c}
        \toprule
        \textbf{Method} &  \textit{PJE} $\downarrow$ & \textit{pr.PJE} $\downarrow$ & \textit{V2V} $\downarrow$ & \textit{pr.V2V} $\downarrow$\\
        \midrule
        No Contact  & 180.2 & 74.0 & 183.3 & 65.2 \\
        \prox~\cite{hassan2019resolving} & 170.9 &	\textbf{72.3}  &	174.0	& 63.4  \\
        All Contact & 175.4 &	73.4 & 176.3 &	\textbf{64.0} \\
		Predicted Contact & \textbf{171.3} & 73.6 & \textbf{172.3} & 64.9 \\    
        \midrule
		GT Contact & 161.9 & 71.8 & 163.0  & 63.3 \\    
        \bottomrule 
     \end{tabular}
}
\caption{
            Contact-driven human pose and shape (HPS) estimation -- results on PROX's ``quantitative set''. 
            ``Predicted Contact''   refers to the contact label predicted by our \hot contact detector and 
            ``GT Contact''          is the ground-truth contact label. 
            ``PROX''                refers to use of PROX's manually annotated contact vertices. 
            ``PJE''                 refers to the Per-Joint Error, 
            ``pr''                  is Procrustes alignment, and 
            ``V2V''                 is the Vertex-to-Vertex error.
}
\label{tab:supp_contact_hps}
\end{table}

\section{HOT for 3D Contact Estimation}
\label{sec:supp_3d_contact}

In the main paper, we show that our \hot dataset facilitates dense \threeD contact estimation on the human body from an image~\cite{huang2022capturing}, by helping such models generalize better to in-the-wild images. 
Below we report how we generate the pseudo ground-truth for \threeD contact using \twoD \hot annotations, and discuss more experimental details.

\begin{figure}[ht!]
	\centering
	\includegraphics[width=1.0\linewidth,trim={0cm 0cm 0cm 0cm},clip]{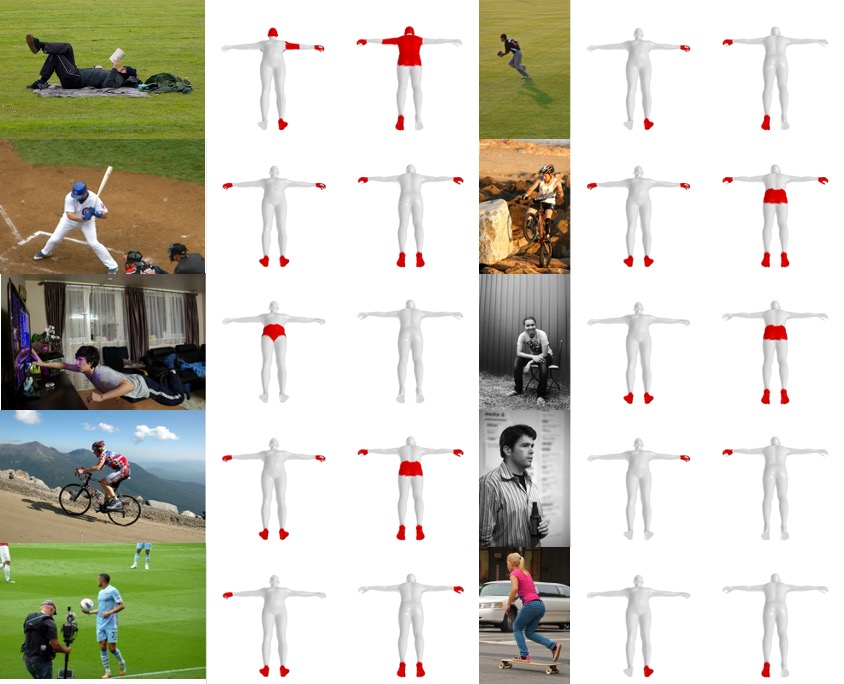}
    \vspace{-0.2em}
	{\small \hspace{0.2cm}  Image \hspace{0.9cm}   F-V \hspace{0.65cm} B-V \hspace{0.3cm}  Image \hspace{0.35cm}   F-V \hspace{0.65cm} B-V \hfill}\\
	\caption{
	            Examples of the pseudo ground-truth 3D contact generated from \setAnnot , \ie , \mbox{HOT-pGT}. 
	            F-V represents front-view and B-V represents back-view. 
	}
	\label{fig:supp_contact3d}
\end{figure}

\qheading{Pseudo ground-truth generation}
For \setAnnot, we annotate (see \cref{sec:supp_contact_anno}) contact areas as 2D polygons in images and the body part that is involved in contact (see part segmentation in \cref{sec:supp_human_part}). 
For the annotated body part, for this experiment we consider all its vertices (see \cref{fig:supp_human_part}) as contact vertices. 
The only exception is the hands and feet; we only consider the vertices on the inner palm and the sole of foot to capture the most common contact in daily life. 
The above results in a coarse pseudo ground-truth 3D contact map on the human body; for examples see \cref{fig:supp_contact3d}.
We denote the pseudo ground-truth 3D contact for \setAnnot as ``\mbox{HOT-pGT}''.

\begin{figure}[ht!]
\vspace{-3mm}
\begin{subfigure}[t]{\linewidth}
\includegraphics[width=\linewidth,trim={0cm 0cm 0cm 0cm},clip]{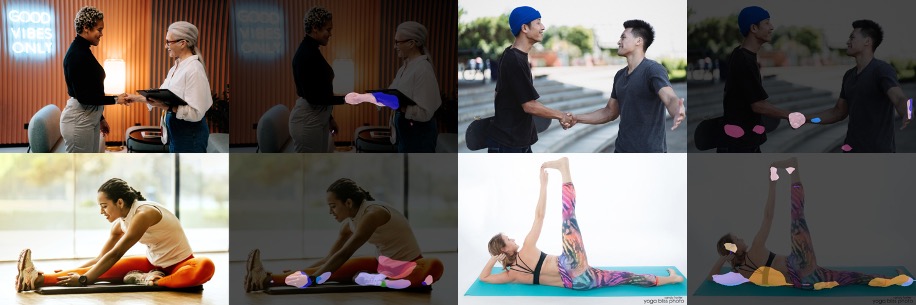}
\end{subfigure}%
\caption{Qualitative results of testing our model on self-contact and human-human contact.}
\label{fig:supp_selfcontact}
\end{figure}
\qheading{Experimental details}  
The recent RICH dataset and BSTRO model \cite{huang2022capturing} focus on dense \threeD contact estimation on the human body from an image.
To show the usefulness of our \hot dataset for this task, we employ the BSTRO model and extend its training dataset RICH with \mbox{HOT-pGT}. 
When training on RICH and \mbox{HOT-pGT}, we combine all the images from the training set of RICH and HOT, following their original training/validation/testing split.
For faster convergence, we use the pre-trained model of BSTRO and fine-tune on the combination of RICH and \mbox{HOT-pGT} for $20$ epochs. The learning rate is set to be $0.0001$ and the the batch size is set to be 32. 
The rest of the network architecture and hyperparameters are the same as original BSTRO training~\cite{huang2022capturing}. 
We compare with the original BSTRO model, which is trained only on RICH.
Each model is evaluated on the test set with the best performer from the validation set.

\section{Contact Detection Applications}
\label{sec:supp_downstream_application}

Contact detection is important for applications in many domains such as AR/VR, activity recognition, affordance detection, fine-grained human-object interaction detection (beyond bounding boxes), \threeD human pose estimation and populating scenes with interacting avatars. 
Here we showcase several examples in \cref{fig:downstream}. 
For instance, one possible future direction is to extend the triplet definition of \ac{hoi} $<$human/action/object$>$ by adding contact as $<$human-part/contact-area/object$>$, which supports finer-grained \ac{hoi} reasoning. 
Another application is detecting in videos the areas that people contact, and guiding human cleaners (AR) or robots with heatmaps for sanitization or contamination prevention. 

We also test our human-object detector on images with self-contact and human-human contact; see some qualitative results in \cref{fig:supp_selfcontact}. 
Although our model was not designed for such interaction scenarios, sometimes it can produce meaningful results, and sometimes it expectedly fails; this is a challenging and open problem.
How to effectively combine different contacts and build a general-purpose contact detector would be interesting future work.

\section{Use of Existing Assets}
\label{sec:supp_assets}

Our dataset \hot collects image data from \prox \cite{hassan2019resolving}, \vcoco~\cite{gupta2015visual}, \hake~\cite{li2020pastanet} and \wnp~\cite{wu2015watch}. 
\prox is licensed under the terms of the Software Copyright License for non-commercial scientific research purposes. 
\vcoco is licensed under the terms of the CC-BY 4.0 License and \hake is licensed under the terms of the MIT License. 

\end{document}